\documentclass[11pt, a4paper, copyright, gdm]{google}

\usepackage[numbers, sort&compress, square]{natbib}
\usepackage{amsmath}
\usepackage{amssymb}
\usepackage{amsfonts}
\usepackage{graphicx}
\usepackage{booktabs}
\usepackage{multirow}
\usepackage{array}
\usepackage{tabularx}
\usepackage{xcolor}
\usepackage{tikz}
\usetikzlibrary{
  positioning, arrows.meta, calc, fit, backgrounds, shapes.geometric
}
\usepackage[most]{tcolorbox}
\usepackage{hyperref}


\uselogo{gdm}

\title{Breaking the high-degree barrier in Lehmer's problem with a
human--AI discovery loop}

\author[1, *]{Yuri Chervonyi}
\author[1,2, *]{Sergei Gukov}
\author[1]{Thang Luong}

\affil[1]{Google DeepMind}
\affil[2]{California Institute of Technology}
\affil[*]{Equal contribution.\protect\\Work done at Google DeepMind.}
\correspondingauthor{cyuri@google.com, sgukov@google.com}

\begin{abstract}
While artificial intelligence has made rapid progress in mathematical reasoning,
many open problems depend on structure that is revealed only through large-scale
computational exploration and analytical synthesis.
One such example is Lehmer's problem from 1933. It asks whether the Mahler measure of an integer polynomial, which measures how far its roots stray from
the unit circle, can be made arbitrarily small.
Previously, constructing irreducible polynomials with integer coefficients
and small Mahler measure has been carried out systematically only for even
degrees $d \le 180$. At higher degrees, Boyd's analytical one-parameter family
frequently factors, skipping nearly $50\%$ of target degrees and leaving no
general construction for arbitrary $d$.
Here, we address the high-degree
regime through a human--AI discovery loop, in which we extend
the conversational prompting with a fast numerical pipeline,
an exact symbolic verifier, and large-scale distributed search.
The AI agent prototypes numerical and symbolic solvers locally,
launches parallel searches across distributed clusters and works with
the human researcher to extract and verify mathematical hypotheses.
Through this interactive loop, we derive a five-parameter construction,
which we call the Generalized Boyd algorithm. It eliminates degree-skipping
and constructs irreducible polynomials with Mahler measure strictly below
Boyd's limit point $\alpha_1 \approx 1.255434$ for every target degree.
Evaluated across all even degrees up to $2000$, our approach improves upon
six previously published values and holds $99.6\%$ of all best known records.
With this work we release the verified dataset of polynomials with minimal
Mahler measure across all studied even degrees.
\end{abstract}

\begin{document}

\maketitle

\setcounter{tocdepth}{1}
{\makeatletter
\let\origaddvspace\addvspace
\renewcommand{\addvspace}[1]{\origaddvspace{0.2em}}
\makeatother
\tableofcontents}
\vspace{1em}

\section{Introduction}
\label{sec:introduction}

Artificial intelligence (AI) has moved, within a decade, from a tool for
recognizing patterns in data to an active participant in scientific
discovery~\cite{wang2023scientific}. Deep networks now predict protein
structures at near-experimental accuracy~\cite{jumper2021highly}, propose
hundreds of thousands of previously unknown stable crystals that autonomous
laboratories have begun to synthesize without human
intervention~\cite{merchant2023scaling,szymanski2023autonomous}, control the
plasma in a tokamak~\cite{degrave2022magnetic}, forecast global weather with
skill rivalling that of operational physical models~\cite{lam2023learning},
place the components of a microchip~\cite{mirhoseini2021graph}, and flag
antibiotic candidates in chemical libraries far too large for exhaustive
screening~\cite{stokes2020deep}. Reinforcement-learning agents descended from
game-playing systems~\cite{silver2016mastering} have discovered faster
matrix-multiplication algorithms~\cite{fawzi2022discovering}. In each of these
cases the decisive ingredient was not the model alone but a \emph{workflow}: a
division of labor in which the machine explores spaces too large or too
tedious for people, and people supply the problem, the priors and the judgement
about what counts as a result.

Increasingly, that division of labor is becoming interactive, with large
language models (LLMs) wired to laboratory instruments and to one another.
Coscientist combined an LLM with information retrieval, code execution and
laboratory automation to plan and perform chemical syntheses on robotic
hardware under researcher-specified objectives~\cite{boiko2023autonomous}. The
Virtual Lab coupled a team of LLM agents with feedback from a single human
scientist to build a nanobody-design pipeline whose products showed
experimentally validated binding to SARS-CoV-2
variants~\cite{swanson2025virtual}, and multi-agent systems have generated and
ranked biomedical hypotheses, some of which were subsequently confirmed at the
bench~\cite{gottweis2025coscientist}. Across these examples, researchers
contribute objectives, domain knowledge and experimental judgement, while the
AI system carries out the computational and procedural work needed to turn a
question into a testable candidate.

Mathematics is a particularly sharp test of such workflows, because here the
final product must be exact rather than merely accurate. AI has already
contributed to mathematical discovery in several ways: supervised models have
surfaced correlations that mathematicians then turned into conjectures and
proofs~\cite{hughes2020neural,davies2021advancing,shehper2024what}; reinforcement learning and
transformers have produced counterexamples and extremal constructions in
combinatorics~\cite{wagner2021constructions,charton2024patternboost};
LLM-driven program evolution has improved bounds on long-standing open
problems~\cite{romera2024mathematical,novikov2025alphaevolve,georgiev2025mathematical};
and neurosymbolic and formal systems have reached medal-level performance at
the International Mathematical
Olympiad~\cite{trinh2024solving,chervonyi2025gold,hubert2025olympiad}. With
respect to the role of the human, however, these approaches often presuppose that the computational infrastructure needed to
evaluate a candidate already exists and is cheap. For many of the most
interesting problems it does not. A conjecture may be testable in principle,
but only if one first builds a fast numerical pipeline, an exact symbolic
verifier and a distributed search that together run at a scale no single
workstation reaches; and the asymptotic regime in which the conjecture becomes
decisive is often precisely the regime that off-the-shelf tools cannot enter.
What has been missing is a workflow in which the AI system does not merely
propose or score candidates but engineers the search and verification
infrastructure itself, runs it at cloud scale, and analyses the returned data
interactively with the researcher so that the search space can be reformulated
as understanding improves.

Interactive human--AI workflows of this kind are therefore essential for
problems that lie beyond the reach of any single existing approach. By
embedding human insight inside scalable, verification-aware discovery loops,
they promise to accelerate progress in mathematics, the physical sciences and
biology without sacrificing rigor, opening regimes that have remained
computationally inaccessible and converting numerical results into analytical
understanding that neither humans nor AI could reach alone.

Here, we develop such a workflow for a problem that exhibits exactly this combination of
computational and conceptual difficulty: Lehmer's problem~\cite{lehmer1933factorization}, open for nine decades, asks whether systems constrained by exact integer
structure can exhibit arbitrarily small positive complexity, or whether
a universal gap separates zero from every positive rate. For a concrete class of deterministic dynamical systems, this is
literally a question about entropy: can their orbit complexity grow
exponentially, but arbitrarily slowly, even as the number of degrees of
freedom increases? The answer would link a fundamental constraint on
polynomial roots to quantitative laws of information growth, arithmetic,
and geometry.

To explain this in more detail, consider a state described by \(d\) real coordinates, each defined modulo one. The state space is a \(d\)-dimensional torus: a hypercube whose opposite faces are identified. Let an integer matrix \(A\) of determinant \(\pm1\) update the state by
\[
 T_A: x \mapsto Ax\pmod{\mathbb{Z}^d}.
\]
Integer entries make the rule consistent with the identifications, and the determinant condition makes it reversible. The states remain continuous; it is the update rule, not the state space, that is constrained by arithmetic.

Topological entropy measures how rapidly the number of distinguishable orbit histories increases as their duration grows and observational resolution becomes finer. If that number grows at exponential rate \(h\),
the information needed to distinguish histories grows at rate \(h/\log2\) bits per iteration. For the systems above, a classical exact formula \cite{Lind1982,LindSchmidtWard} is
\[
 h_{\mathrm{top}} (T_A)
   =\sum_{|\lambda_j|>1}\log|\lambda_j|
   =m (P),
\]
where \(\lambda_j\) are the eigenvalues, $P(x) = \det(xI-A)$ is the characteristic polynomial, and $m(P)$ will be defined shortly in \eqref{eq:mahler_definition} --- it will be the main character of our story. Thus, the arithmetic quantity is also a dynamical information rate. For example, Arnold's cat map $(x,y)\mapsto(2x+y,\,x+y)$ on the torus is the textbook example of a chaotic system, which corresponds to \(A=\left(\begin{smallmatrix}2&1\\1&1\end{smallmatrix}\right)\). 
Its characteristic polynomial \(x^2-3x+1\) has expanding eigenvalue
\((3+\sqrt5)/2\), giving topological entropy
\(\log((3+\sqrt5)/2)\approx0.9624\). 
Lehmer's conjecture asserts that chaos is \emph{quantised} for such systems: a linear integer map of a torus of any dimension is either not chaotic at all, or has entropy at least a universal constant, conjecturally attained in dimension $10$.

The original motivation for Lehmer's conjecture comes from number theory~\cite{lehmer1933factorization}.
Consider a polynomial whose coefficients are ordinary integers, $P(x) = \sum_{k=0}^d a_k x^k
\in \mathbb{Z}[x]$, $a_d \ne 0$. Factor it over the complex numbers, $P(x) = a_d \prod_{j=1}^d (x - z_j)$, and look at its roots $z_j$. Some lie inside the unit circle, some on it, some outside. Multiply together the absolute values of the roots that lie \emph{outside} the circle (and the leading coefficient $a_d$). The resulting number, now known as the \emph{Mahler measure},
\begin{equation}
  M(P) = e^{m(P)} := |a_d| \prod_{j=1}^d \max(1, |z_j|).
  \label{eq:mahler_definition}
\end{equation}
is a single quantity that summarises how far the polynomial's roots stray from the unit circle. It is always at least $1$, and a theorem of Kronecker from 1857 says it equals $1$ exactly when every root is a root of unity, i.e.\ a point $e^{2\pi i k/n}$ on the circle with rational angle. Lehmer's question (1933) asks:
\begin{quote}
\emph{Can the Mahler measure of an integer polynomial be larger than $1$, but only by an arbitrarily small amount?}
\end{quote}
Lehmer's conjecture says no: there is a definite gap above $1$. The smallest value anyone has ever found is the measure of the polynomial
\begin{equation}
  L(x) = x^{10} + x^9 - x^7 - x^6 - x^5 - x^4 - x^3 + x + 1, \quad M(L)
  \approx 1.1762808182599175,
  \label{eq:lehmer_poly}
\end{equation}
which Lehmer found by hand-guided search. Ninety years and many computer searches later, it remains the record.
It is proved optimal for all $d \le
54$~\cite{boyd1989reciprocal,mossinghoff1998list,mossinghoff2008minimal} and
conjectured optimal across all degrees~\cite{lehmer1933factorization}.

Lehmer is the right person to have asked this. He was a pioneer of computing who built mechanical and photoelectric sieves for factoring integers, and his question arose from a practical goal: he wanted integer sequences that grow as slowly as possible, so as to find large primes efficiently. The growth rate of his sequences is exactly the Mahler measure, and Lehmer's question asks whether there is a universal speed limit from below.

The question can be explained to an undergraduate, yet the best general result, Dobrowolski's bound of 1979~\cite{dobrowolski1979question}, is still essentially the state of the art: a lower bound that tends to $1$ as the degree grows\footnote{Originally proved with leading constant $1 - \varepsilon$, and later
refined to $2 - \varepsilon$ by Cantor and
Straus~\cite{cantor1982conjecture}.}
\begin{equation}
  M(P) \ge 1 + (2 - \varepsilon) \left( \frac{\log \log d}{\log d} \right)^3
  \quad (d \to \infty).
\label{eq:dobrowolski}
\end{equation}
No one has been able to remove this decay in more than forty-five years.

\begin{figure}[t!]
  \centering
  \includegraphics[width=\textwidth]{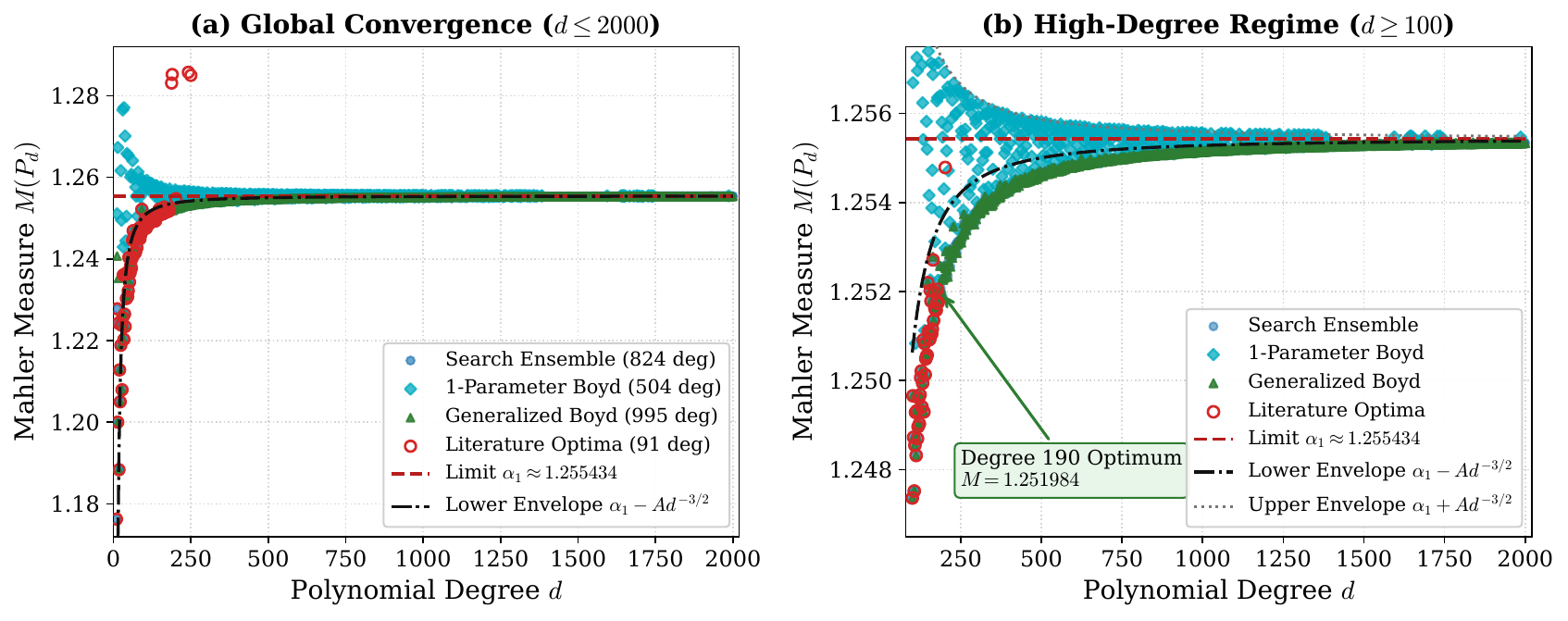}
  \caption{\textbf{Global convergence of univariate Mahler measures at high
  degree $d$ to Boyd's limit $\alpha_1$.} \textbf{(a)} Comparison of published
  Literature Optima (91 degrees, red open circles), the computational Search
  Ensemble discovered by reinforcement learning, evolutionary search, and
  periodic defect tiling (824 degrees, blue circles), the 1-Parameter Boyd
  family (504 degrees, cyan diamonds), and the Generalized Boyd family (995
  degrees, green triangles) against the limit $\alpha_1 \approx 1.255434$ (red
  dashed line) and lower envelope $\alpha_1 - A d^{-3/2}$ (black dash-dot
  curve). \textbf{(b)} High-degree regime ($d \ge 100$), showing how the
  Generalized Boyd family follows the lower envelope $\alpha_1 - A d^{-3/2}$
  strictly below $\alpha_1$, whereas the 1-Parameter Boyd family oscillates
  between the lower and upper envelopes $\alpha_1 \pm A d^{-3/2}$.}
  \label{fig:mahler_envelope}
\end{figure}

This is precisely what the present paper is aiming to address, with the help of AI. Indeed, Lehmer's problem has a rare two-sided character. The strong form (``$M(L)$ is the minimum'') would be refuted by a \emph{single, finite, machine-checkable} polynomial, and exhaustive searches have so far excluded counterexamples of degree up to $54$~\cite{mossinghoff2008minimal}. Refuting the conjecture itself, by contrast, would require an infinite family that escapes every known construction. This makes it a natural target for the new generation of AI-assisted search and program-synthesis methods~\cite{romera2024mathematical,charton2024patternboost,novikov2025alphaevolve}, and a sharp test of their limits~\cite{shehper2024what}.
While prior results show a decay of the lower bound a la \eqref{eq:dobrowolski}, our AI-assisted bounds \emph{increase} (not decrease!) as $d$ grows, cf. Figure~\ref{fig:mahler_envelope}.

\paragraph{Why the High Degree is the Difficulty.} For each fixed degree, a gap follows from elementary coefficient bounds.
Indeed, bounded degree and bounded Mahler measure imply that only finitely
many integer polynomials are possible. The conjecture concerns the failure
of that compactness argument when the degree grows. It asks whether an
arithmetic quantity can stay separated from zero even while its ambient
dimension becomes arbitrarily large.

To understand
how actual minimal Mahler measures behave for large degrees,
Boyd~\cite{boyd1981speculations} and Lawton~\cite{lawton1983problem} proved
that one-variable specializations $F(x, x^n)$ of two-variable polynomials $F(X,
Y)$ converge, as $n \to \infty$, to the two-variable Mahler measure $M(F(X,
Y))$, which forms a closed set of limit points. In 1981,
Boyd~\cite{boyd1981speculations} identified the smallest known limit point as
the two-variable Mahler measure $\alpha_1 = M(F) \approx 1.255433866$ of
$F(X, Y) = X(X+1)Y^2 + (X^2+X+1)Y + (X+1)$.

Despite Boyd's analysis of the one-parameter sequence $F(x, x^n)$, determining
the minimum Mahler measure over \emph{irreducible non-cyclotomic} polynomials
at each fixed even\footnote{Basic symmetry constraints help to narrow down the search for small Mahler measures. In 1971, Smyth~\cite{smyth1971product} proved that for any non-reciprocal polynomial $P(x) \ne \pm x^d P(1/x)$, the Mahler measure satisfies:
\begin{equation}
  M(P) \ge M(x^3 - x - 1) \approx 1.324717957244746.
  \label{eq:smyth_bound}
\end{equation}
Consequently, any polynomial satisfying $1 < M(P) < 1.324718$ must be reciprocal ($a_k = \pm a_{d-k}$). Since any reciprocal polynomial of odd degree vanishes at $x = \pm 1$ and is therefore reducible for $d > 1$, all odd degrees are immediately resolved, restricting the search for irreducible polynomials with minimal Mahler measure exclusively to even degrees $d = 2m$.} degree $d$,
\begin{equation}
  M_{\min}(d) = \min \left\{ M(P) : P \in \mathbb{Z}[x] \text{ irreducible,
  non-cyclotomic, } \deg(P) = d \right\},
\end{equation}
has been systematically studied only for $d \le 180$ via searches by
Boyd~\cite{boyd1981speculations,boyd1989reciprocal},
Mossinghoff~\cite{mossinghoff1998list}, Boyd and
Mossinghoff~\cite{boyd2005small}, and Mossinghoff, Rhin, and
Wu~\cite{mossinghoff2008minimal} (with isolated benchmarks up to $d \le 250$,
comprising 91 published literature degrees).
Extending explicit constructions and unconstrained searches to arbitrary large
even degrees $d > 180$ confronts three obstacles:
\begin{enumerate}
  \item \textbf{Exponential Search Space}: In
    unconstrained fixed-degree search over reciprocal polynomials with integer
    coefficients in $\{-2, -1, 0, 1, 2\}$, the number of candidate polynomials
    scales as $5^{d/2}$ (exceeding $10^{349}$ candidates at $d = 1000$).
  \item \textbf{Unconstrained Testing of Boyd's Limit-Point Conjecture}:
    Without exploring the high-degree coefficient space directly via general
    optimization (such as reinforcement learning and evolutionary search), it
    remained untested whether unconstrained high-degree search ($d \gg 200$)
    would uncover any smaller limit points below $\alpha_1$ or whether
    independent search trajectories always collapse onto specializations of
    $F(X, Y)$.
  \item \textbf{Cyclotomic Reducibility and Degree-Skipping in 1-Parameter
    Families}: For a given exponent $n$, a one-parameter specialization such as
    $F(x, x^n)$ is frequently reducible over $\mathbb{Q}$, factoring into a
    product of cyclotomic\footnote{The $n$-th \emph{cyclotomic polynomial}
$\Phi_n(x) = \prod_{1 \le k \le n, \, \gcd(k,n)=1} (x - e^{2\pi i k/n}) \in
\mathbb{Z}[x]$ is the unique monic irreducible integer polynomial whose roots
are the primitive $n$-th roots of unity on the unit circle $|z| = 1$.} polynomials $\Phi(x) = \prod_m \Phi_m(x)$ of varying
    degree $\delta(n)$ times an irreducible factor of shifted degree $d =
    \deg(F(x, x^n)) - \delta(n)$. Because $\delta(n)$ jumps discontinuously
    with the divisors of $n$, dividing out cyclotomic factors skips nearly
    $50\%$ of target degrees $d$, while $M(F(x, x^n))$ oscillates both
    above and below
    $\alpha_1$.
\end{enumerate}

To overcome these computational bottlenecks, we developed a fast numerical
evaluation and rational irreducibility verification pipeline
(Section~\ref{sec:efficient-search}) and combined it with reinforcement
learning, evolutionary search, and priority search to navigate the high-degree
coefficient space (Section~\ref{sec:search_algorithms}). Deploying these search
methods across degrees up to $d = 2000$ revealed that while low degrees admit
isolated minima down to $1.17628$ ($d=10$) and $1.18837$ ($d=18$), for all $d
\ge 30$ the empirical minima across every independent search method rise
asymptotically and accumulate along the horizontal limit $\alpha_1 \approx
1.255434$ from below (Figure~\ref{fig:mahler_envelope}). Inspecting the
algebraic structure of the polynomials along this barrier in turn enabled us to
resolve the cyclotomic degree-skipping obstacle through a generalized
multi-parameter analytical construction
(Section~\ref{sec:analytic_resolution}).

\paragraph{Significance of the Problem.} Lehmer's problem belongs to a class of problems that specialists across several fields regard as fundamental, and whose solution would be front-page news within mathematics. Its standing is documented in several places beyond specialist papers.
For example, Wolfram MathWorld explicitly includes ``Lehmer's Mahler
measure problem'' in its \emph{Unsolved Problems} collection, which also
includes the Riemann hypothesis, Goldbach's conjecture, and \(P\) versus
\(NP\) \cite{MathWorldProblems}.
Lehmer's conjecture is included in Wikipedia's widely consulted \emph{List of unsolved problems in mathematics}~\cite{wikiunsolved}, and it was the subject of the ``WHAT IS\ldots?'' column of the \emph{Notices of the American Mathematical Society}~\cite{hironaka2009what}, a series reserved for concepts of broad current interest. It is the organizing theme of dedicated surveys~\cite{smyth2008mahler,smyth2015seventy,ghate2001arithmetic} and of recent monographs~\cite{mckee2021around,brunault2020many}.

On its face Lehmer's question is a narrow statement about the distribution of algebraic integers near the unit circle. Its importance comes from the fact that the Mahler measure $M(P)$ or its logarithmic variant $m(P) = \log M(P)$ appears, often in disguise, as a fundamental invariant in many fields:

\begin{itemize}[leftmargin=2em]
\item in \textbf{Diophantine geometry} it is $d$ times the absolute logarithmic Weil height of a root, so Lehmer's conjecture is a sharp, degree-sensitive lower bound for heights~\cite{bombieri2006heights,smyth2008mahler};
\item in \textbf{ergodic theory} it is the topological (and measure-theoretic) entropy of the associated toral automorphism~\cite{yuzvinskii1967calculation}, and in several variables the entropy of algebraic $\mathbb{Z}^n$-actions~\cite{lind1990ergodic,schmidt1995dynamical,everest1999heights};
\item in \textbf{complex dynamics} the entropy of an automorphism of a compact complex surface is the logarithm of a Salem number~\cite{cantat2001dynamique,mcmullen2002dynamics}, and the smallest possible positive value is $m(L) = \log M(L)$~\cite{mcmullen2007dynamics,mcmullen2016automorphisms};
\item in \textbf{geometric group theory} $M(L)$ is the minimal spectral radius greater than one of a Coxeter element~\cite{mcmullen2002coxeter}, and also the minimal Mahler measure greater than one of an integer symmetric matrix~\cite{mckee2012integer};
\item in \textbf{low-dimensional topology} $m(\Delta_K) = \log M (\Delta_K)$ is the exponential growth rate of torsion in the homology of cyclic branched covers of a knot $K$~\cite{riley1990growth,gonzalezacuna1991cyclic,silver2002mahler,le2014homology}, and $L$ is the Alexander polynomial of the $(-2,3,7)$ pretzel knot~\cite{hironaka2001lehmer};
\item in \textbf{operator algebras and $L^2$-invariants} $M(f)$ is the Fuglede--Kadison determinant of $f$ acting on $\ell^2(\mathbb{Z}^n)$~\cite{luck2002l2}, and Lehmer's problem extends to arbitrary groups~\cite{luck2022lehmer};
\item in \textbf{hyperbolic geometry} Salem numbers are exponentials of lengths of closed geodesics on arithmetic hyperbolic orbifolds~\cite{neumann1992arithmetic,maclachlan2003arithmetic,emery2019salem}, and a counterexample to Lehmer of Salem type would yield arbitrarily short geodesics;
\item in \textbf{arithmetic geometry} multivariable Mahler measures are periods~\cite{kontsevich2001periods}, and in many proven and conjectural cases they equal special values of $L$-functions and of regulators~\cite{smyth1981mahler,deninger1997deligne,boyd1998mahler,rodriguez1999modular,brunault2020many}, as predicted by Beilinson's conjectures~\cite{beilinson1985higher};
\item in \textbf{mathematical physics} they appear as free energies of dimer models~\cite{kenyon2006dimers} and as hyperbolic volumes governing the asymptotics of quantum invariants~\cite{boyd2002mahler,gukov2005three}.
\end{itemize}

A solution of Lehmer's problem, in either direction, would therefore propagate into each of these fields as a uniform \emph{gap} or as the existence of arbitrarily small positive invariants.

\paragraph{If Lehmer's Conjecture is True.} Our results provide evidence in favor of Lehmer's conjecture. Among multiple applications, within \emph{number theory} a proof would give the sharp, degree-sensitive height bound $h(\alpha)\ge\kappa/\deg\alpha$ for every algebraic number that is not a root of unity. It would also answer Salem's question by bounding Salem numbers away from $1$~\cite{ghate2001arithmetic,smyth2015seventy}. It would also extend, with the same constant, to Mahler measures in any number of variables, and hence to the special values of $L$-functions, regulators and periods that these measures represent~\cite{boyd1998mahler,deninger1997deligne,kontsevich2001periods,brunault2020many}. Effective results built on lower bounds for heights would shed the $(\log\log d/\log d)^3$ losses inherited from Dobrowolski-type bounds~\cite{dobrowolski1979question,voutier1996effective}. These include linear forms in logarithms~\cite{matveev2000explicit}, estimates for regulators of number fields~\cite{friedman1989analytic} and certified computation with integer polynomials~\cite{mignotte1974inequality,lenstra1982factoring}. The bound for the house, by contrast, would not improve, since Dimitrov's constant already exceeds $m(L) =  \log M(L)$~\cite{dimitrov2019proof}. The elliptic, abelian and dynamical analogues~\cite{laurent1983minoration,masser1989counting,amoroso1999probleme,david2000minoration,call1993canonical} would not follow formally. Every unconditional result in those settings has, however, been obtained by transporting techniques from the multiplicative group, so a method that breaks the Dobrowolski barrier would very likely transfer to them.

Across the rest of mathematics, the conjecture becomes a family of uniform ``quantisation'' statements. In \emph{dynamics}, every toral or solenoidal automorphism, and every algebraic $\mathbb{Z}^n$-action, with positive entropy would have entropy at least $m(L) = \log M(L)$~\cite{yuzvinskii1967calculation,lind1990ergodic,lind2005lehmer,schmidt1995dynamical}. A pseudo-Anosov map of genus $g$ with non-cyclotomic homological action would satisfy $\log\lambda\ge(m(L))/g$, a homological counterpart of Penner's bound~\cite{penner1991bounds,leininger2004groups,mcmullen2000polynomial,farb2011small}. In \emph{topology}, torsion in the homology of abelian covers of any knot or link would grow either subexponentially or at rate at least $m(L)$~\cite{riley1990growth,gonzalezacuna1991cyclic,silver2002mahler,le2014homology}. Fuglede--Kadison determinants of group-ring elements over free abelian groups would be bounded away from $1$, settling the abelian case of L\"uck's problem~\cite{luck2022lehmer}. Non-abelian groups behave differently. The L\"uck--Schick formula gives, for the figure-eight knot group, an element of determinant $\exp(\mathrm{vol} (4_1)/6\pi)=1.1137\ldots<M(L)$~\cite{luck1999l2,luck2002l2}, and torsion growth in non-abelian congruence towers is governed by hyperbolic volume instead~\cite{bergeron2013asymptotic}. In \emph{hyperbolic geometry}, closed geodesics on arithmetic hyperbolic $2$- and $3$-orbifolds would have length bounded below by a universal constant (the Short Geodesic Conjecture), making closed arithmetic hyperbolic $3$-manifolds uniformly thick~\cite{neumann1992arithmetic,maclachlan2003arithmetic,emery2019salem}. In \emph{mathematical physics}, free energies of periodic dimer models with integer weights, and Mahler measures of $A$-polynomials, would obey the same gap~\cite{kenyon2006dimers,boyd2002mahler,gukov2005three}. Finally, the gap is already a theorem for automorphisms of complex surfaces, Coxeter groups and integer symmetric matrices~\cite{mcmullen2002coxeter,mcmullen2007dynamics,mcmullen2016automorphisms,mckee2012integer}. There the strong form would reveal $M(L)$ and the $E_{10}$ lattice as universal extremal objects of arithmetic rather than artifacts of positivity.

\paragraph{If Lehmer's Conjecture is False} A counterexample would be a sequence of irreducible non-cyclotomic polynomials $P_k \in \mathbb{Z} [x]$ with $M(P_k)=1+\varepsilon_k \to 1$, and known results already force such polynomials to be very unusual.
The strong form is different: it would be refuted by a single polynomial with Mahler measure in the interval $(1,M(L))$, which must have degree greater than $54$~\cite{mossinghoff2008minimal}.
The consequences would reach the same areas as a proof. In \emph{number theory}, the height bound would fail and Dobrowolski-type bounds would be optimal up to the exponent. If the counterexamples were Salem numbers, these would accumulate at $1$, so the set of Salem and Pisot numbers would not be closed, contrary to a conjecture of Boyd~\cite{ghate2001arithmetic}. In \emph{dynamics}, there would be toral automorphisms of arbitrarily small positive entropy~\cite{lind2005lehmer}, sharply separating integer matrices from complex surfaces, where the gap is a theorem. In \emph{topology and operator algebras}, Fuglede--Kadison determinants are unchanged under induction from a subgroup, so the Lehmer constant in L\"uck's sense would equal $1$ for every group containing an element of infinite order~\cite{luck2002l2,luck2022lehmer}. If the counterexamples satisfied $P_k(1)=\pm1$, they would be Alexander polynomials of knots whose branched covers have torsion growing exponentially at arbitrarily small rates. In \emph{hyperbolic geometry}, counterexamples of Salem type would give arithmetic hyperbolic surfaces with arbitrarily short closed geodesics, so arithmetic groups would not be uniformly discrete~\cite{neumann1992arithmetic,emery2019salem}. In \emph{arithmetic geometry}, the elliptic and dynamical Lehmer conjectures would lose their main heuristic support~\cite{laurent1983minoration,call1993canonical}. In \emph{mathematical physics}, it would become natural to ask whether such polynomials can be spectral curves of dimer models, which are Harnack curves, with arbitrarily small free energy~\cite{kenyon2006dimers}.

\paragraph{Summary of Results.} In this work, we address the construction and characterization of
minimal-Mahler polynomials for arbitrary high degrees $d$ --- and
systematically benchmark our constructions across all even degrees up to $d =
2000$ (Table~\ref{tab:ensembles_overview} and Figure~\ref{fig:mahler_envelope})
--- through four main contributions:

\begin{table}[t!]
  \centering
  \caption{\textbf{Summary of polynomial families} evaluated on the high-degree
  benchmark ($d \le 2000$).}
  \label{tab:ensembles_overview}
  \small
  \renewcommand{\arraystretch}{1.18}
  \begin{tabularx}{\textwidth}{@{}lrX@{}}
    \toprule
    \textbf{Polynomial Family} & \textbf{\# of $d$'s} &
    \textbf{Key Empirical \& Theoretical Properties} \\
    \midrule
    Literature Optima~\cite{mossinghoff2008minimal} & 91 & Known minima for
    $d \le 250$ (systematic coverage up to $d \le 180$). \\
    Search Ensemble & 824 &
    RL, evolutionary, and defect-tiling optima; matches 85 and beats 6 of the
    91 Literature Optima. \\
    1-Parameter Boyd & 504 &
    One-parameter specialization; skips nearly $50\%$ of target degrees; beats
    the Search Ensemble on 118 degrees. \\
    Generalized Boyd & 995 &
    Applies to any even $d \ge 12$; verified irreducible with $M(P_d) <
    \alpha_1$ for all $d \in [12, 2000]$ ($99.6\%$ of records). \\
    \bottomrule
  \end{tabularx}
\end{table}

\begin{enumerate}
  \item \textbf{Interactive Human--AI Collaborative Workflow
    (Section~\ref{sec:agent_workflow})}: Moving beyond static prompting or
    closed-loop code mutation over a fixed evaluator, our workflow pairs
    Gemini\footnote{Throughout this work, we used internal versions of Gemini,
    including Gemini~4 Argon~\cite{gemini4argon2026}.}
    with Google Antigravity~\cite{antigravity2026} --- an agentic development
    environment that connects the model to a persistent code workspace, local
    terminal and symbolic algebra tools, and Google Cloud cluster job
    orchestration --- running rapid prototyping and analysis locally while
    launching large-scale distributed sweeps across Google Cloud CPU and GPU
    clusters.
  \item \textbf{Numerical Search Algorithms for Small Mahler Measure
    (Section~\ref{sec:search_algorithms})}: After resolving the core
    computational bottleneck of fast exact irreducibility verification,
    we combine reinforcement learning, evolutionary search,
    and periodic defect tilings across reciprocal polynomials to establish
    upper bounds across 824 even degrees in the Search Ensemble, matching 85
    and strictly improving upon 6 of the 91 published Literature Optima,
    including $d = 178$ within the systematically searched $d \le 180$ regime:
\begin{align*}
  P_{178} & (x) = x^{178} + x^{177} + x^{176} - x^{174} - 2x^{173} - 2x^{172} - x^{171} + x^{169} + x^{168} + x^{167} + x^{163} + x^{162} \\
  &+ x^{161} - x^{159} - 2x^{158} - 2x^{157} - x^{156} + x^{154} + x^{153} + x^{152} + x^{148} + x^{147} + x^{146} - x^{144} \\
  &- 2x^{143} - 2x^{142} - x^{141} + x^{139} + x^{138} + x^{137} + x^{133} + x^{132} + x^{131} - x^{129} - 2x^{128} - 2x^{127} \\
  &- x^{126} + x^{124} + x^{123} + x^{122} + x^{118} + x^{117} + x^{116} - x^{114} - 2x^{113} - 2x^{112} - x^{111} + x^{109} \\
  &+ x^{108} + x^{107} + x^{103} + x^{102} + x^{101} - x^{99} - 2x^{98} - 2x^{97} - x^{96} + x^{94} + x^{93} + x^{92} \\
  &- x^{89} + x^{86} + x^{85} + x^{84} - x^{82} - 2x^{81} - 2x^{80} - x^{79} + x^{77} + x^{76} + x^{75} + x^{71} \\
  &+ x^{70} + x^{69} - x^{67} - 2x^{66} - 2x^{65} - x^{64} + x^{62} + x^{61} + x^{60} + x^{56} + x^{55} + x^{54} \\
  &- x^{52} - 2x^{51} - 2x^{50} - x^{49} + x^{47} + x^{46} + x^{45} + x^{41} + x^{40} + x^{39} - x^{37} - 2x^{36} \\
  &- 2x^{35} - x^{34} + x^{32} + x^{31} + x^{30} + x^{26} + x^{25} + x^{24} - x^{22} - 2x^{21} - 2x^{20} - x^{19} \\
  &+ x^{17} + x^{16} + x^{15} + x^{11} + x^{10} + x^{9} - x^{7} - 2x^{6} - 2x^{5} - x^{4} + x^{2} + x + 1
\end{align*}
  \item \textbf{The Generalized Boyd Algorithm
    (Section~\ref{sec:analytic_resolution})}: By jointly analyzing the
    unconstrained search optima and clearing their cyclotomic factors, we show
    that $95.5\%$ of the search optima for $d \ge 30$ collapse onto
    specializations of Boyd's two-variable
    polynomial~\cite{boyd1981speculations,boyd1998mahler},
    $F(X, Y) = X(X + 1)Y^2 + (X^2 + X + 1)Y + (X + 1)$, with limit point
    $\alpha_1 = M(F) \approx 1.25543387$. Combining the RL-discovered phase
    shifts with Boyd's asymptotic expansion yields a five-parameter
    \emph{Generalized Boyd algorithm} that eliminates cyclotomic
    degree-skipping by solving directly for the exponents required to hit any
    target even degree $d$, thereby generating low-Mahler candidates for
    arbitrarily large $d$. Benchmarked across all 995 even degrees $d \in [12,
    2000]$, the Generalized Boyd family constructs irreducible polynomials
    strictly below $\alpha_1$ for every degree ($\max_{d \in [12, 2000]}
    M(P_d^*) = 1.25535826 < \alpha_1$ at $d = 1932$), holding $99.6\%$ of all
    best-known records (including $100\%$ of all degrees $d \ge 166$).
  \item \textbf{Benchmark Dataset and Physical Interpretation 
    (Sections~\ref{sec:spectral_distributions}--\ref{sec:dataset})}: By
    interpreting $\log M(P)$ as the electrostatic potential of a
    two-dimensional Coulomb gas of point charges and counting roots in exact
    rational arithmetic over $\mathbb{Q}$, we show across all 824 benchmark
    polynomials that a scale-invariant fraction of $77.0\%\text{--}88.0\%$
    (mean $82.51\%$) of their roots lie strictly on the unit circle
    $|z|=1$ while the remaining off-circle roots lie in a narrow layer of width
    $\mathcal{O}(1/d)$ around the unit circle.
    Finally, we release the complete verified benchmark dataset across all 996
    even degrees $d \in [10, 2000]$ in both expanded and compact
    cyclotomic-quotient forms.
\end{enumerate}

\section{Human--AI Discovery Loop}
\label{sec:agent_workflow}

In applications of AI to mathematics, several paradigms have emerged. In the ``guided intuition'' approach, a supervised model
trained on precomputed invariants highlights correlations that mathematicians
then turn into conjectures and proofs, as in the discovery of new relations
between different types of knot invariants and in representation
theory~\cite{hughes2020neural,davies2021advancing}; a similar interplay between learned policies
and human analysis has been brought to bear on the Andrews--Curtis
conjecture~\cite{shehper2024what}. In the ``construction'' approach,
reinforcement learning or transformer models sample discrete objects against a
fixed reward, yielding counterexamples and extremal configurations in graph
theory and combinatorics~\cite{wagner2021constructions,charton2024patternboost}.
Closed-loop program-evolution systems replace the sampled object with a sampled
program: FunSearch improved bounds on the cap-set problem and on online bin
packing~\cite{romera2024mathematical}, and AlphaEvolve has extended the method
to matrix multiplication and to a sweep of dozens of open optimization problems
across analysis, combinatorics, geometry and number theory, in several cases
matching or improving the best known
constructions~\cite{novikov2025alphaevolve,georgiev2025mathematical}. Finally,
neurosymbolic and formal systems couple neural proposers with symbolic engines
or proof assistants: AlphaGeometry and AlphaProof reached medal-level
performance at the International Mathematical
Olympiad~\cite{trinh2024solving,chervonyi2025gold,hubert2025olympiad}, and
agentic workflows have steered local constraint solvers towards new
combinatorial designs~\cite{xia2026agentic}. The importance of these efforts
lies less in any single theorem than in the demonstration that machine search
and human mathematical reasoning can be composed into a loop, with each side
correcting the blind spots of the other.

\begin{figure}[t!]
  \centering
  \resizebox{\textwidth}{!}{%
  \begin{tikzpicture}[
    node distance=1.55cm and 2.25cm,
    box/.style={
      rectangle, rounded corners=4pt, draw=black!80, thick, fill=gray!8,
      text width=4.50cm, minimum height=1.55cm, align=center, font=\small
    },
    agentbox/.style={
      rectangle, rounded corners=4pt, draw=blue!75!black, very thick,
      fill=blue!6, text width=4.50cm, minimum height=1.55cm, align=center,
      font=\small
    },
    enginebox/.style={
      rectangle, rounded corners=4pt, draw=green!55!black, thick, fill=green!6,
      text width=4.50cm, minimum height=1.55cm, align=center, font=\small
    },
    arrow/.style={-{Stealth[length=2.5mm, width=1.8mm]}, thick, draw=black!75},
    lbl/.style={font=\footnotesize, align=center, inner sep=3pt}
  ]
    \node[box] (human) {
      \textbf{Human Expert}\\
      Formulates Problems,\\
      Conjectures \& Symmetries
    };
    \node[agentbox, right=2.25cm of human] (agent) {
      \textbf{AI Agent}\\
      (Gemini + Antigravity)\\
      Reasoning, Code \& Tools
    };
    \node[enginebox, right=2.25cm of agent] (compute) {
      \textbf{Google Cloud (GCP)}\\
      CPU \& GPU Clusters\\
      Large-Scale Experiments
    };

    \node[box, below=1.55cm of agent] (dataset) {
      \textbf{Knowledge Base}\\
      Validated Examples\\
      \& Key Properties
    };
    \node[box] (deconv) at (dataset.center -| human.center) {
      \textbf{Analytical Synthesis}\\
      Pattern Compression\\
      \& Closed-Form Models
    };
    \node[enginebox] (verify) at (dataset.center -| compute.center) {
      \textbf{Automated Verification}\\
      Symbolic, Numerical\\
      \& Consistency Checks
    };

    \draw[arrow] ([yshift=6pt]human.east) --
      node[above, lbl] {Priors \&\\Guidance} ([yshift=6pt]agent.west);
    \draw[arrow] ([yshift=-6pt]agent.west) --
      node[below, lbl] {Findings \&\\Insights} ([yshift=-6pt]human.east);
    \draw[arrow] (agent.east) --
      node[above, lbl] {Launch\\Cloud Jobs} (compute.west);
    \draw[arrow] (compute.south) --
      node[right, lbl] {Candidate\\Results} (verify.north);
    \draw[arrow] (verify.west) --
      node[above, lbl] {Verified\\Results} (dataset.east);
    \draw[arrow] (dataset.north) --
      node[right, lbl] {Validation\\Loop} (agent.south);
    \draw[arrow] (dataset.west) --
      node[above, lbl] {Empirical\\Patterns} (deconv.east);
    \draw[arrow] (deconv.north) --
      node[left, lbl] {Analytical\\Model} (human.south);
  \end{tikzpicture}%
  }
  \caption{\textbf{Interactive human--AI discovery loop built on Gemini and
  Google Antigravity.} A continuous validation loop connects local agentic
  reasoning and tool-building with large-scale cloud experiments and automated
  verification on Google Cloud (CPU and GPU clusters), while human--AI
  collaboration distills empirical patterns into general analytical models.}
  \label{fig:agent_architecture}
\end{figure}
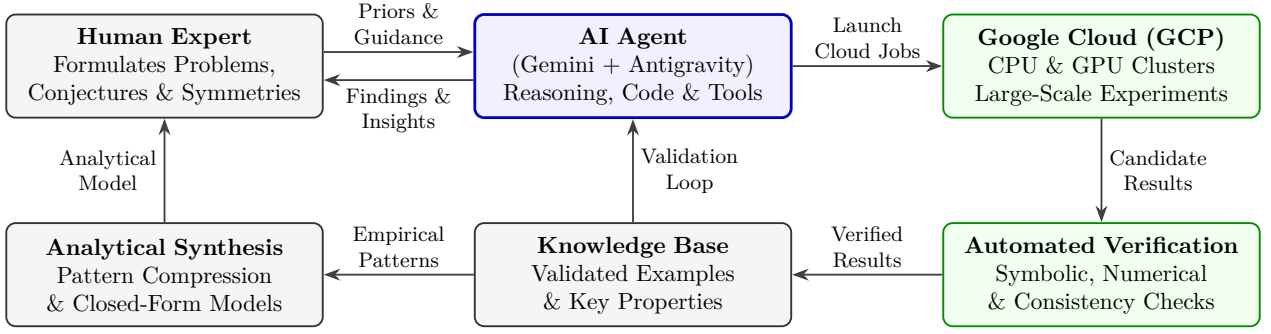

A central methodological contribution of this study is our human--AI
collaborative discovery workflow built on Gemini~\cite{gemini4argon2026} and
Google Antigravity~\cite{antigravity2026} (illustrated in
Figure~\ref{fig:agent_architecture}). Architecturally, Google Antigravity is an
agentic development environment and execution harness that connects the Gemini
reasoning model to a persistent workspace equipped with file-editing tools,
local shell execution, symbolic computer-algebra tools (Mathematica),
background task monitors, and cluster job launchers for Google
Cloud CPU and GPU pools. Unlike closed-loop evolutionary code-mutation systems
that operate as isolated black-box evaluators over fixed scoring functions, our
interactive human-in-the-loop workflow integrates mathematical reasoning,
algorithm design, and high-performance software engineering into a unified
discovery loop.

In high-degree Mahler measure minimization, neither brute force nor standard
reinforcement learning can succeed in isolation unless two foundational
computational bottlenecks are solved first: rapid numerical evaluation of
polynomial roots across millions of candidate polynomials, and exact
verification of irreducibility over $\mathbb{Q}$. Throughout the investigation,
our collaborative workflow operated across three integrated pillars:
(i)~decoupling search workers from asynchronous four-step exact verification
over $\mathbb{Q}$;
(ii)~translating domain priors into complementary search algorithms scaled
across cloud compute; and (iii)~synthesizing empirical optima into
closed-form analytical constructions via cyclotomic factoring.

\paragraph{Comparison with Existing AI Discovery Approaches.}
Table~\ref{tab:ai_math_comparison} contrasts the four paradigms surveyed above
with our discovery loop, which advances interactive neurosymbolic collaboration
into a fifth paradigm: \emph{distributed cloud experimentation, scalable
algorithm engineering, and symbolic hypothesis verification}. Whereas recent
agentic neurosymbolic studies such as~\cite{xia2026agentic} couple
human steering with local constraint solvers and simulated annealing in
single-node sessions, high-degree polynomial optimization up to $d = 2000$
cannot be solved by local scripts or off-the-shelf solvers alone. Our framework
distinguishes itself through three general capabilities:
(a)~\emph{two-tier local-and-cloud distributed orchestration}, reserving local
execution for rapid prototyping, sanity checks, and interactive data analysis
while autonomously launching and monitoring large-scale distributed numerical
and symbolic sweeps across Google Cloud CPU and GPU clusters,
(b)~\emph{decoupled numerical search and symbolic verification}, where the
agent separates fast numerical candidate generation from asynchronous exact
irreducibility verification over $\mathbb{Q}$ so search workers never stall on
symbolic algebra, and
(c)~\emph{iterative search-space reformulation and analytical synthesis},
co-evolving candidate parameterizations with the human researcher from
unconstrained discrete vectors to structured families before symbolically
factoring empirical optima to derive closed-form analytical constructions
(Sections~\ref{sec:search_algorithms}--\ref{sec:analytic_resolution}).

\begin{table}[t!]
  \centering
  \caption{\textbf{Comparison of Human--AI Mathematical Discovery Paradigms.}}
  \label{tab:ai_math_comparison}
  \small
  \begin{tabularx}{\textwidth}{@{}
    >{\raggedright\arraybackslash}p{3.1cm}
    >{\raggedright\arraybackslash}p{3.3cm}
    >{\raggedright\arraybackslash}p{3.2cm}
    >{\raggedright\arraybackslash}X @{}}
    \toprule
    \textbf{Discovery Paradigm} & \textbf{Representative Works} &
    \textbf{AI Role in Loop} & \textbf{Primary Output \& Limitation} \\
    \midrule
    Standalone RL / Neural Sampling &
    Wagner~\cite{wagner2021constructions},
    AlphaTensor~\cite{fawzi2022discovering},
    PatternBoost~\cite{charton2024patternboost} &
    Fixed-reward policy optimization &
    Isolated finite counterexamples; no closed-form symbolic law. \\
    \addlinespace[4pt]
    ML-Guided Human Intuition &
    Davies et al.~\cite{davies2021advancing},
    Shehper et al.~\cite{shehper2024what} &
    Feature attribution \& pattern guidance &
    Human-formulated conjecture; AI does not run iterative experiments. \\
    \addlinespace[4pt]
    Closed-Loop Code Evolution &
    FunSearch~\cite{romera2024mathematical},
    AlphaEvolve~\cite{novikov2025alphaevolve} &
    Automated heuristic mutation &
    Evolved search scripts against a static pre-written evaluator. \\
    \addlinespace[4pt]
    Formal Proving \& Local Neurosymbolic &
    AlphaGeometry~\cite{trinh2024solving,chervonyi2025gold},
    Co-Scientist~\cite{gottweis2025coscientist},
    Xia et al.~\cite{xia2026agentic} &
    Deductive search \& local solver steering &
    Formal proofs or bounds derived from single-node / small-$n$ solver
    exploration; no distributed cloud scale or asynchronous symbolic
    verification at scale. \\
    \addlinespace[4pt]
    \textbf{Interactive Cloud Experiment--Hypothesis--Verification} &
    \textbf{This Work (Gemini + Antigravity)} &
    \textbf{Co-designs domain solvers, scales cloud sweeps, and runs exact
    symbolic checks} &
    \textbf{Synthesizes large-scale numerical optima into closed-form
    analytical constructions.} \\
    \bottomrule
  \end{tabularx}
\end{table}

\section{Numerical Search Algorithms for Small Mahler Measure}
\label{sec:search_algorithms}

In this work, we are interested in finding polynomials with integer coefficients and a minimal Mahler measure in a given degree $d$. Such polynomials have been constructed explicitly only for degrees up to 250. Within this range, a systematic numerical search was done for degrees up to 180~\cite{boyd1981speculations,boyd2005small,mossinghoff2008minimal}, while polynomials with $d \le 54$ are proven to have the minimal Mahler measure~\cite{mossinghoff2008minimal}. Exhaustive search for higher degrees is not feasible because the number of polynomial candidates grows exponentially with degree, which is why we focus on sparse search methods.

\subsection{Efficient search}
\label{sec:efficient-search}

We start by creating a leaderboard that records for each degree the polynomial with the smallest known Mahler measure. To make the search more efficient and scalable, we incorporate several improvements. First, by Smyth's theorem~\cite{smyth1971product}, any polynomial with $M(P) <
1.324718$ must be reciprocal ($P(x) = x^d P(1/x)$) of even degree $d = 2m$, so we reduce the number of integer coefficients by half. Second, we notice that low-Mahler measure candidates are frequently reducible\footnote{A polynomial
$P(x)$ is \emph{reducible} if it factors as a product $P(x) =
A(x)B(x)$ of lower-degree polynomials.} or arise from lower degrees as $P(x) = Q(x^k)$. In order to maintain the search workers' speed, we pair them with asynchronous verification workers. Search workers filter out duplicates and stream candidates with $1.01 < M(P) < 1.3$ without pausing the search, while the verification worker performs the following steps:
\begin{enumerate}
    \item \textbf{Deflate polynomial.} Let $k$ be the greatest common divisor of all powers
    of $x$ appearing in $P(x)$. When $k > 1$, $P(x) = Q(x^k)$ has the same Mahler
    measure $M(Q(x^k)) = M(Q(x))$ as a lower-degree polynomial $Q(x)$, so we     replace $P(x)$ by $Q(x)$ and record it at its true degree $d/k$.
    \item \textbf{Check irreducibility.} Check whether the deflated polynomial is reducible via \texttt{IrreduciblePolynomialQ} in Mathematica, and disregard reducible ones.
    \item \textbf{Compute Mahler measure.} Find the complex roots $z_1, \dots, z_d$ numerically via \texttt{numpy.roots}, evaluate $M(P) = \prod_{j=1}^d \max(1, |z_j|)$, and discard cyclotomic polynomials\footnote{Recall that an irreducible integer polynomial is \emph{cyclotomic} if all of its roots are on the unit circle $|z_j| = 1$, and therefore its Mahler measure is $1$.}.
    \item \textbf{Add to leaderboard.} If the computed Mahler measure is smaller than recorded on the leaderboard we overwrite the existing entry, otherwise we disregard the candidate.
\end{enumerate}

\subsection{Search Methods}
\label{sec:search_methods}

After setting up the leaderboard and verification protocol, we develop five
complementary search strategies. We begin with simple sparse baselines such as
direct search and nearest neighbors, followed by reinforcement learning,
evolutionary search, and repeating short coefficient blocks with a central
defect. The methods are described below, and the results are summarized in
Table~\ref{tab:search_harvest}.

\paragraph{Direct Sparse Search.}
We start with the simplest possible baseline: searching directly over
reciprocal polynomials with only $5$ or $7$ non-zero terms,
\begin{equation}
  P(x) = x^d + 1 + c_0 x^{d/2}
  + \sum_{j=1}^{s} c_j \big(x^{k_j} + x^{d - k_j}\big),
  \label{eq:direct_sparse}
\end{equation}
where $s \in \{1, 2\}$, $c_0 \in \{-1, 0, +1\}$, $1 \le k_1 < \dots < k_s <
d/2$ are the non-zero powers, and $c_j \in \{-1, +1\}$ are their signs.

\paragraph{Priority Search.}
To find dense low-Mahler polynomials that have many non-zero terms, we next use
a simple nearest-neighbor search. Recalling that cyclotomic
polynomials $P_0$ have all their roots on the unit circle (and therefore
have $M(P_0) = 1$), we observe that after changing a few integer
coefficients by $+1$ or $-1$ most roots still stay on the unit circle
and only some move slightly outside. This process produces non-cyclotomic
polynomials with $M(P)$ close to 1. To explore this idea, we use a priority
queue: initialize it with a cyclotomic polynomial $P_0$, pop the highest-scoring
candidate, perturb one coefficient at a time within $[-2, 2]$, and push
the unvisited neighbors back into the queue with score
\begin{equation}
  \mathrm{score}(P) = \frac{1}{|M(P) - 1|} \cdot 0.1^{\max(0,\,
  N_{\mathrm{off}} - 4)} \cdot 0.95^{\|P - P_0\|_1} \cdot \eta,
  \label{eq:priority_score}
\end{equation}
where $N_{\mathrm{off}}$ is the number of roots off the unit circle, $\|P -
P_0\|_1$ counts the $\pm 1$ edits from $P_0$, and $\eta \sim
\mathrm{Uniform}[0.85, 1.15]$ injects random exploration noise.

\paragraph{Reinforcement Learning.}
While nearest neighbors perturbations work well for small degrees,
single edits quickly become very inefficient as degree grows.
A natural way to address this is to apply Reinforcement learning,
which showed great success in solving problems with a high dimensional
state/action space, as was pioneered in AlphaGo~\cite{silver2016mastering}.
Similarly to the priority search, we want to produce sparse polynomials;
to do so we define a policy by assigning each
coefficient $a_k$ an independent probability distribution
$p_{k,-1}, p_{k,0}, p_{k,1}$ over $\{-1, 0, 1\}$.
These distributions are initialized with high probability at $0$, so that
sampled polynomials are sparse. At each iteration of the Cross-Entropy
Method (CEM), we sample polynomials
from $p_{k, v}$, compute their Mahler measure $M(P)$, and score each
non-cyclotomic candidate ($M(P) > 1$) using
\begin{equation}
  R(P) = -\log M(P) - 0.8\max(0,\, M(P) - 1.6)
  + \lambda_{\mathrm{uc}} f_{\mathrm{uc}}(P) +
  \begin{cases}
    10, & M(P) < 1.1763, \\
    3, & 1.1763 \le M(P) < 1.3, \\
    0, & \text{otherwise},
  \end{cases}
  \label{eq:cem_reward}
\end{equation}
where $f_{\mathrm{uc}}(P)$ is the fraction of roots on the unit circle (while
cyclotomic polynomials with $M(P) = 1$ receive $R(P) = -10$). We then select
the top-scoring polynomials in the batch, count the fraction $f_{k,v}$ of them
that have $a_k = v$ for each $v \in \{-1, 0, +1\}$, and move the sampling
probabilities toward those fractions:
\begin{equation}
  p_{k,v} \leftarrow (1 - \alpha)\, p_{k,v} + \alpha\, f_{k,v},
  \qquad \alpha = 0.6.
  \label{eq:cem_update}
\end{equation}
Finally, we refine the best polynomials in the batch by changing one
coefficient at a time in $\{-1, 0, +1\}$ whenever doing so lowers $M(P)$.

\paragraph{Evolutionary Search.}
Because our RL approach averages all top candidates into a single probability table,
distinct polynomial families at different degrees can blur together. To keep a
diverse pool of candidates across degrees, we maintain a population of
low-Mahler polynomials evolved directly through algebraic operations: symmetric
splicing ($x^a P_1(x) \pm P_2(x)$ and single-point crossover), dividing out
cyclotomic factors, and symmetric single-coefficient steps $P(x) \pm (x^{d-k} +
x^k)$. Without diversity control, tournament selection is quickly overrun by
multiples of Lehmer's degree-$10$ polynomial and other low-degree optima ($d
\le 18$). We prevent this by ranking candidates with the penalized cost
\begin{equation}
  \mathrm{cost}(P) = M(P) \cdot \big(1 + 0.1\max(0,\, n_Q - 5)\big) +
  \begin{cases}
    1, & \text{if } P \text{ is divisible by a } d \le 18
        \text{ optimum}, \\
    0, & \text{otherwise},
  \end{cases}
  \label{eq:evo_cost}
\end{equation}
where $n_Q$ counts how many candidates in the population share the same
deflated polynomial $Q(x)$.

\paragraph{Periodic Tiling with Defects.}
Inspecting the dense polynomials found by RL revealed a striking pattern: their
coefficients repeat a short block of length $L$ from both ends, interrupted
only by a symmetric defect around the middle power $x^{d/2}$. We search
directly over this family:
\begin{equation}
  P(x) = \sum_{q=0}^{K-1} (\pm 1)^q \Big(x^{qL} B(x)
  + x^{d - qL} B(x^{-1})\Big) + x^{d/2} \Delta(x),
  \label{eq:periodic_tiling_defect_main}
\end{equation}
where $B(x) = \sum_{r=0}^{L-1} b_r x^r$ ($b_0 = 1$, $b_r \in \{-1, 0, +1\}$) is
a repeating block of length $L$, $K = \lfloor d / (2L) \rfloor$ counts how many
blocks fit from each end, and $\Delta(x)$ is a short reciprocal polynomial in
$x$ and $x^{-1}$ with coefficients in $\{-1, 0, +1\}$ filling the remaining gap
around $x^{d/2}$. Because we only choose the short block $B(x)$ and central
defect $\Delta(x)$ rather than all $d/2$ coefficients, this method scales
cleanly to $d = 2000$ and motivates our analytical construction in
Section~\ref{sec:analytic_resolution}.

\subsection{Numerical Search Results}
\label{sec:numerical_search_results}

\begin{figure}[!ht]
  \centering
  \includegraphics[width=\textwidth]{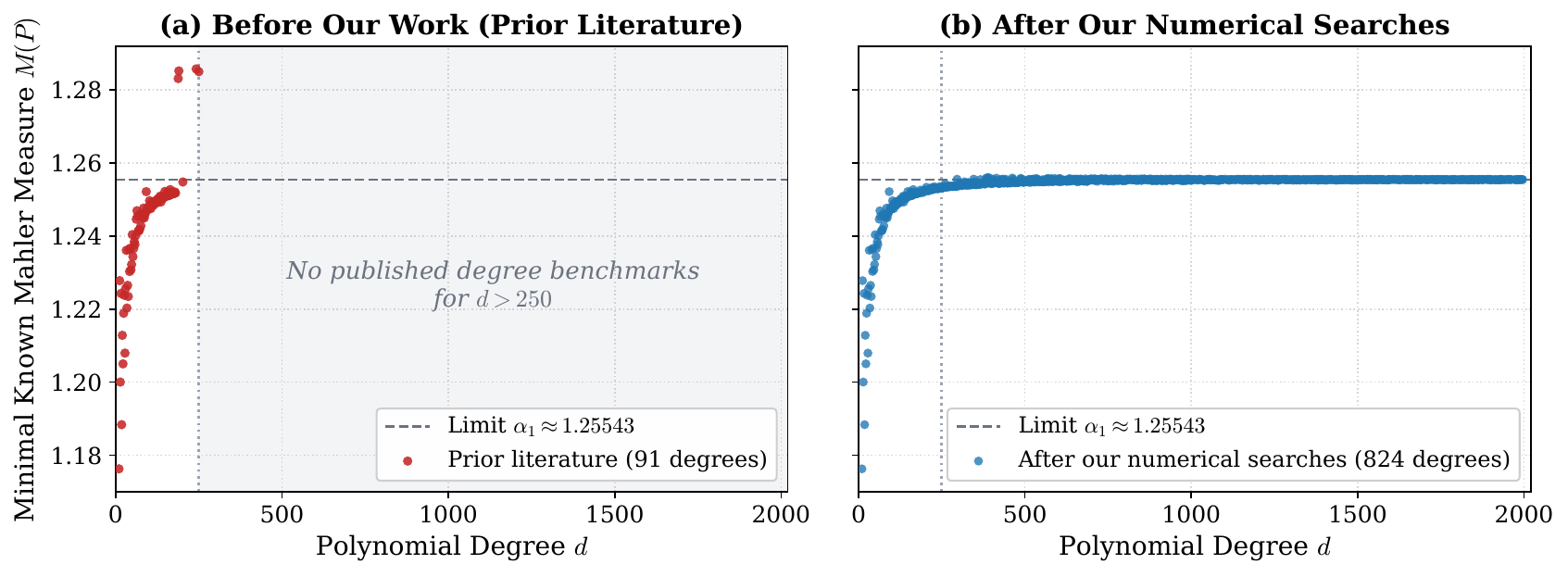}
  \caption{\textbf{Minimal known Mahler measure.} \textbf{(a)}~Prior literature 
  covers 91 even degrees up to $d \le 250$
  ~\cite{boyd1981speculations,boyd2005small,mossinghoff2008minimal}.
  \textbf{(b)}~Our numerical search methods establish 824 even degree optima
  across $d \le 2000$, strictly improving 6 prior literature bounds at $d \le
  250$ and establishing the upper bounds across 733 additional degrees.}
  \label{fig:numerical_search_before_after}
\end{figure}

Our results are summarized in Figure~\ref{fig:numerical_search_before_after} where we compare the best literature bounds prior to this work with those obtained by our numerical searches. A breakdown across algorithms is presented in Table~\ref{tab:search_harvest}. Across the 91 degrees published in the literature ($d \le 250$), our search matches 85 and achieves strict improvements on 6 degrees: $d = 178$ (within the $d \le 180$ regime where intense systematic searches were previously conducted) and $d = 188$, $190$, $202$,
$242$, and $250$ (Table~\ref{tab:lit_improvements}). We also establish upper bounds across 733 previously unexplored even degrees up to $d = 2000$. This raises a question: why all numerical methods converge to the same asymptote, and what underlying principle governs it?

\begin{table}[!ht]
  \centering
  \caption{\textbf{Our leaderboard entries ($d \le 2000$) by search method.}}
  \label{tab:search_harvest}
  \small
  \begin{tabular*}{\textwidth}{@{\extracolsep{\fill}}l r r l@{}}
    \toprule
    \textbf{Algorithm} & \textbf{Wins} &
    \textbf{Share} & \textbf{Degrees} \\
    \midrule
    Periodic Tiling + Defect & 447 & $54.2\%$ &
    $d \in [10, 1846]$ \\
    RL Cross-Entropy Policy & 186 & $22.6\%$ &
    $d \in [14, 800]$ \\
    Direct Sparse Search & 111 & $13.5\%$ &
    $d \in [44, 1996]$ \\
    Priority Search & 57 & $6.9\%$ &
    $d \in [20, 290]$ \\
    Evolutionary Search & 23 & $2.8\%$ &
    $d \in [12, 300]$ \\
    \midrule
    \textbf{Total} & \textbf{824} & \textbf{100.0\%} &
    $d \in [10, 2000]$ \\
    \bottomrule
  \end{tabular*}
\end{table}

\begin{table}[!ht]
  \centering
  \caption{\textbf{Strict improvements over literature for $d \le 250$.}}
  \label{tab:lit_improvements}
  \small
  \begin{tabular*}{\textwidth}{@{\extracolsep{\fill}}r r r r@{}}
    \toprule
    \textbf{Degree $d$} & \textbf{Literature $M_{\mathrm{lit}}$} &
    \textbf{New Bound $M(P_d^*)$} & \textbf{Improvement $\Delta M$} \\
    \midrule
    $178$ & $1.25204881$ & $1.25162839$ & $-0.00042$ \\
    $188$ & $1.28309200$ & $1.25207822$ & $-0.03101$ \\
    $190$ & $1.28518461$ & $1.25198376$ & $-0.03320$ \\
    $202$ & $1.25478722$ & $1.25253255$ & $-0.00225$ \\
    $242$ & $1.28571755$ & $1.25305628$ & $-0.03266$ \\
    $250$ & $1.28497453$ & $1.25316520$ & $-0.03181$ \\
    \bottomrule
  \end{tabular*}
\end{table}

\section{The Generalized Boyd Algorithm}
\label{sec:analytic_resolution}

The empirical convergence of five independent computational search methods
toward the same asymptote $\alpha_1 \approx 1.255434$ from below
(see Figure~\ref{fig:numerical_search_before_after}(b)) indicates that high
degree polynomials with minimal Mahler measure share a common algebraic origin.
Here we show that multiplying these high degree polynomials by
suitable cyclotomic polynomials collapses them onto one-variable
specializations of Boyd's two-variable polynomial
$F(X, Y)$~\cite{boyd1981speculations}.
Based on this observation we formulate a five-parameter algorithm, which we
call the generalized Boyd algorithm, that eliminates cyclotomic
degree-skipping for any target even degree $d \ge 12$ and yields verified
irreducible polynomials with Mahler measure strictly below $\alpha_1$ across
all tested degrees $d \le 2000$.

\begin{table}[t!]
  \centering
  \caption{\textbf{Comparative performance of our search algorithms,
  One-parameter Boyd family $Q_n(x)$, and the Generalized Boyd algorithm
  $Q_{a,b,k,s_0,s_1}(x)$} evaluated on the benchmark ($d \le 2000$), alongside
  several explicit examples.}
  \label{tab:boyd_improvements}
  \small
  \begin{tabular*}{\textwidth}{@{\extracolsep{\fill}}l r r r c@{}}
    \toprule
    \textbf{Family} & \textbf{Degrees Covered} &
    \textbf{Best / Tied Records} & \textbf{Max $M$} &
    \textbf{All $M < \alpha_1$?} \\
    \midrule
    Search Ensemble & 824 & 163 & $1.255965$ & No \\
    One-parameter Boyd & 504 & 19 & $1.311999$ & No \\
    Generalized Boyd & \textbf{995} & \textbf{991 ($99.6\%$)} &
    $\mathbf{1.255358}$ & \textbf{Yes} \\
    \bottomrule
  \end{tabular*}
  \vspace{0.6em}

  \begin{tabular*}{\textwidth}{@{\extracolsep{\fill}}l r l r r l@{}}
    \toprule
    \multicolumn{6}{@{}l}{\textbf{Examples obtained by the Generalized Boyd
    algorithm $Q_{a,b,k,s_0,s_1}(x)$:}} \\
    \midrule
    \textbf{Degree $d$} & \textbf{Prev.\ $M$} &
    \textbf{Source} & \textbf{Gen.\ Boyd $M$} &
    \textbf{$\Delta M$} & \textbf{Tuple $(a, b, k, s_0, s_1)$ \& $\Phi$} \\
    \midrule
    $d = 188$ & $1.283092$ & Literature &
    $1.252078$ & $-0.03101$ & $(11, 86, 0, +1, +1) / \Phi_7$ \\
    $d = 242$ & $1.285718$ & Literature &
    $1.253056$ & $-0.03266$ & $(10, 111, 0, -1, -1)$ ($\delta=0$) \\
    $d = 346$ & $1.255578$ & RL (CEM) &
    $1.253921$ & $-0.00166$ & $(10, 163, 0, +1, +1)$ ($\delta=0$) \\
    $d = 442$ & $1.255552$ & Tiling &
    $1.254495$ & $-0.00106$ & $(14, 207, 0, +1, +1)$ ($\delta=0$) \\
    $d = 526$ & $1.255604$ & Tiling &
    $1.254664$ & $-0.00094$ & $(11, 256, 0, +1, -1) / \Phi_{30}$ \\
    $d = 746$ & $1.255582$ & Tiling &
    $1.254988$ & $-0.00059$ & $(1, 372, 0, +1, -1)$ ($\delta=0$) \\
    $d = 1500$ & $1.255369$ & Tiling &
    $1.255282$ & $-0.00009$ & $(139, 611, 0, -1, -1)$ ($\delta=0$) \\
    \bottomrule
  \end{tabular*}
\end{table}

\subsection{Collapsing Empirical Results via Cyclotomic Factoring}

We start by considering three explicit polynomials found by our algorithms in
Section~\ref{sec:search_algorithms}: $P_{188}(x)$, $P_{600}(x)$, and
$P_{1846}(x)$. Inspecting these high-degree polynomials with hundreds of terms
reveals that they consist of short blocks of coefficients that repeat
periodically from both ends.
As a result, multiplying by a suitable product of cyclotomic polynomials
$\Phi(x) = \prod_m \Phi_m(x)$ cancels all interior repeating blocks while
leaving the Mahler measure unchanged since $M(\Phi) = 1$.

First, in $P_{188}(x)$ we notice eleven repeated blocks
$x^{11 + 7j}(1 - x + x^3 - x^4)$ for $j \in [0, 10]$ from both lower and upper
ends.
Then using the identity
\begin{equation}
    (1 - x)\Phi_7(x) = (1 - x)(1 + x + x^2 + x^3 + x^4 + x^5 + x^6) = 1 - x^7,
\end{equation}
and multiplying the repeated blocks by $\Phi_7(x) = (1 - x^7)/(1 - x)$ we get
\begin{equation}
\Phi_7(x)\sum_{j=0}^{10} x^{11 + 7j}(1 - x + x^3 - x^4) =
\frac{1-x^7}{1-x}\sum_{j=0}^{10} x^{11 + 7j}(1 - x)(1 + x^3) = 
(1 + x^3)x^{11}\sum_{j=0}^{10} (x^{7j} - x^{7(j+1)}).
\end{equation}
Almost all terms inside the sum cancel out leaving only $1 - x^{77}$. So multiplying
the entire $P_{188}(x)$ by $\Phi_7(x)$ gives:
\begin{equation}
  P_{188}(x) \cdot \Phi_7(x) = x^{194} + x^{183} + x^{108} + x^{97} + x^{86}
  + x^{11} + 1.
  \label{eq:deg188_telescope}
\end{equation}
We further notice that the right-hand side is exactly Boyd's two-variable
polynomial~\cite{boyd1981speculations}
\begin{equation}
  F(X, Y) = X(X + 1)Y^2 + (X^2 + X + 1)Y + (X + 1)
  \label{eq:boyd_f_xy}
\end{equation}
evaluated at $X = x^{11}$ and $Y = x^{86}$.

Second, for $P_{600}(x)$ we have terms repeating the sign-alternating 
block $(-1)^j x^{15j}(1 + x^3 + x^5 + x^8)$
for $j = 0, \dots, 19$. Similarly to the previous considered polynomial, 
multiplying $P_{600}(x)$ by 
$(1 + x^{15}) = \Phi_2(x)\Phi_6(x)\Phi_{10}(x)\Phi_{30}(x)$ 
cancels all interior terms leaving:
\begin{equation}
  (1 + x^{15}) P_{600}(x) = (x^5 + 1) \Big[ x(x^3 + 1) y^2 - (x^4 + x^2 + 1)
  y + (x^3 + 1) \Big], \quad y = x^{303},
  \label{eq:deg600_telescope}
\end{equation}
which can be written as $\Phi_{30}(x) P_{600}(x) = F(x, -x^{303})$.

Third, in $P_{1846}(x)$, the interior terms repeat the sign-alternating block
$(-1)^j x^{11 + 6j}(-1 + x - x^2 + x^3)$ for $j = 0, \dots, 149$. Multiplying
$P_{1846}(x)$ by $(1 + x^6) = \Phi_4(x)\Phi_{12}(x)$ cancels consecutive blocks
of opposite sign and collapses the polynomial into $14$ terms:
\begin{equation}
  (1 + x^6) P_{1846}(x) = (x^2 + 1) \Big[ u(u - 1) y^2 - (u^2 - u + 1) y - (u
  - 1) \Big], \quad u = x^{11}, \; y = x^{914},
  \label{eq:deg1846_telescope}
\end{equation}
which can be written as $\Phi_{12}(x) P_{1846}(x) =
F(-x^{11}, -x^{914})$.

Motivated by these examples, we systematically test cyclotomic multipliers
across all best polynomials found in Section~\ref{sec:search_algorithms}.
For each polynomial $P_d(x)$, we searched over products of cyclotomic
polynomials $\Phi(x) = \prod_m \Phi_m(x)$ minimizing the number of non-zero
terms in the product $P_d(x)\Phi(x)$. Remarkably, $95.5\%$ of our dense
polynomials for $d\ge 30$ collapse to sparse ones with only $6$ or $7$ non-zero
terms, and $99.7\%$ of those sparse polynomials arise from Boyd's single
two-variable
polynomial~\eqref{eq:boyd_f_xy} under the four sign choices $(X, Y) \mapsto
(s_0 x^a, s_1 x^b)$ with $s_0, s_1 \in \{-1, +1\}$.\footnote{The four sign
pairs $(s_0, s_1) \in \{-1, +1\}^2$ span all four sign combinations of
seven-term polynomials, while six-term specializations arise from $(x^a +
1)F(-x^a, -x^b)$ via $(x^a + 1)(x^{2a} - x^a + 1) = x^{3a} + 1$.}

Finally, noting again that $M(P_d \cdot \Phi) = M(P_d)$, Boyd's limit
theorem~\cite{boyd1981speculations} implies that as $b/a \to \infty$, the
Mahler measure of $F(s_0 x^a, s_1 x^b)$ converges to the two-variable Mahler
measure $M(F(X, Y)) = \alpha_1 \approx 1.25543387$ (see
Appendix~\ref{app:boyd_limit} for the derivation). This explains why all five
search methods in Section~\ref{sec:search_algorithms} approach the same
asymptote.

\subsection{Five-Parameter Analytical Construction}

To turn this asymptotic theory into an explicit constructive algorithm for any
even degree $d \ge 12$, we study one- and five-parameter families of $F(X, Y)$.

\paragraph{One-Parameter Family.}
From the Degree-600 identity~\eqref{eq:deg600_telescope}, we consider a
one-parameter family that fixes the low exponents and varies $y = x^n$:
\begin{equation}
  Q_n(x) = x(x^3 + 1) x^{2n} - (x^4 + x^2 + 1) x^n + (x^3 + 1).
  \label{eq:qn_1param}
\end{equation}
After dividing out cyclotomic factors $\Phi(x)$ and verifying irreducibility,
$Q_n(x)$ yields irreducible polynomials across $504$ even degrees, strictly
improving upon our numerical search results from
Section~\ref{sec:search_algorithms} across $118$ degrees
(see Table~\ref{tab:boyd_improvements}). However, $Q_n(x)$ skips $375$ even degrees
(such as $d = 188$ and $d = 1276$) due to discrete jumps in the degree of its
cyclotomic factor.\footnote{For example, $Q_{92}(x)$ (of degree $188$) is
divisible by $\Phi_6(x)$ of degree $2$ (giving $d = 186$), whereas $Q_{93}(x)$
(of degree $190$) is divisible by both $\Phi_6(x)$ and $\Phi_{30}(x)$ of total
degree $10$ (jumping to $d = 180$ and skipping $d = 188$).}

\paragraph{Five-Parameter Generalized Boyd Algorithm.}
To eliminate cyclotomic degree-skipping for any target even degree $d \ge 12$,
we propose the five-parameter \emph{Generalized Boyd algorithm}:
\begin{equation}
  \begin{aligned}
    Q_{a,b,k,s_0,s_1}(x) &= x^{a+k}(x^a + s_0)x^{2b}
      + s_1(x^{2a} + s_0 x^a + 1)x^b + s_0(x^a + s_0), \\
    a, b &\in \mathbb{Z}^+, \quad 0 \le k < a, \quad s_0, s_1 \in \{-1, +1\}.
  \end{aligned}
  \label{eq:generalized_boyd_5param}
\end{equation}
Notice that $Q_{a,b,k,s_0,s_1}(x)$ has degree $2b + 2a + k$ (and reduces to
$F(s_0 x^a, s_1 x^b)$ when $k = 0$). After dividing out its cyclotomic
factor $\Phi(x)$ of degree $\delta = \deg\Phi(x)$, the resulting polynomial
$P(x) = Q_{a,b,k,s_0,s_1}(x)/\Phi(x)$ has degree $d = 2b + 2a + k - \delta$.
Therefore, to construct a polynomial of a given target degree $d$, we solve
this degree relation for $b$:
\begin{equation}
  b = \frac{d + \delta - 2a - k}{2}.
  \label{eq:exact_diophantine_b}
\end{equation}
Because $\delta = \deg\Phi(x)$ is itself determined by $(a, b, k, s_0, s_1)$,
$\delta$ in Eq.~\eqref{eq:exact_diophantine_b} acts as a target cyclotomic
degree: for each candidate tuple, we set $b$ via~\eqref{eq:exact_diophantine_b}
and retain the candidate if and only if the
actual cyclotomic factor $\Phi(x)$ of $Q_{a,b,k,s_0,s_1}(x)$ has degree
$\deg\Phi(x) = \delta$, ensuring that $P(x) = Q_{a,b,k,s_0,s_1}(x)/\Phi(x)$ has
exact degree $d$. As we prove in Appendix~\ref{app:cyclotomic_bound}, for $k =
0$ (where all reciprocal polynomials occur) and $\gcd(a, b) = 1$, the
cyclotomic factor $\Phi(x)$ can only consist of divisors from $\{\Phi_7,
\Phi_{12}, \Phi_{14}, \Phi_{15}, \Phi_{30}\}$ and its degree is universally
bounded by $\delta \le 14$ ($\delta \in \{0, 4, 6, 8, 10, 14\}$) for all $d$.
Furthermore, Appendix~\ref{app:self_consistent_existence} proves that a
self-consistent cyclotomic-free solution ($\delta = 0, k = 0$) is guaranteed to
exist for every even degree $d \ge 12$.
To summarize, for each target degree $d$, we vary $(a, k, s_0, s_1)$ and even $\delta
\le 36$, filter for self-consistent tuples with $\deg\Phi(x) = \delta$,
and select the irreducible quotient $Q_{a,b,k,s_0,s_1}(x)/\Phi(x)$ with minimal
Mahler measure.

Evaluating this construction across all $995$ even degrees $d \in [12, 2000]$
produces a verified irreducible polynomial strictly below $\alpha_1 \approx
1.255434$ for every degree, approaching $\alpha_1$ from below at rate
$\mathcal{O}(d^{-3/2})$ (see Appendix~\ref{app:trough_scaling} for the
scaling analysis). Across $d \in [12, 2000]$, the Generalized
Boyd algorithm holds $80.8\%$ of the best-known records at the baseline cutoff
$a \le 20$ and $99.6\%$ across the extended parameter sweep ($a \le \lfloor
(d+\delta)/4 \rfloor$), including every degree $d \ge 166$ (see
Appendix~\ref{app:dilation_saturation} for the parameter cutoff analysis).

\section{Physical Interpretation}
\label{sec:spectral_distributions}

\begin{figure}[t!]
  \centering
  \includegraphics[width=\textwidth]{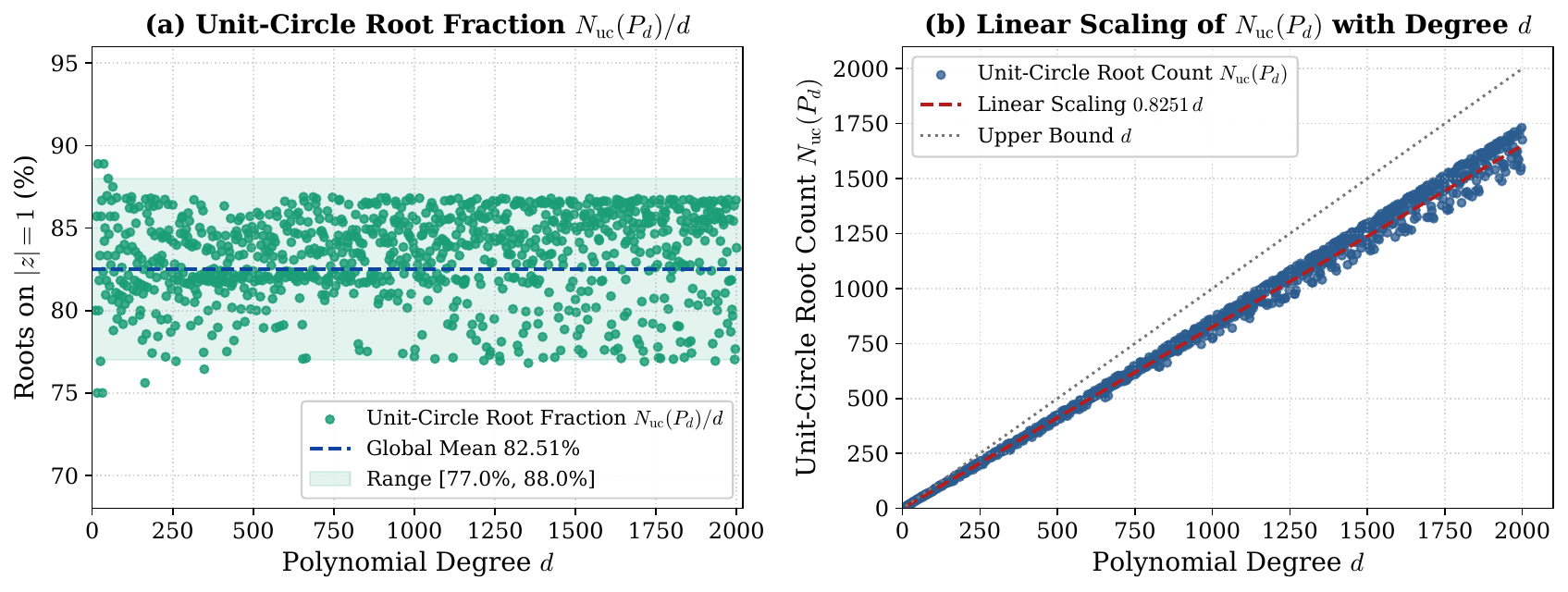}
  \caption{\textbf{Scale invariance of unit-circle root condensation
  established via exact root counting over $\mathbb{Q}$.} \textbf{(a)}
  Unit-circle root percentage $100 \times N_{\mathrm{uc}}(P_d)/d$ across all
  824 optimal polynomials, showing that optimal polynomials remain tightly
  concentrated within $77.0\%\text{--}88.0\%$ (global mean $82.51\%$) on
  $|z|=1$. \textbf{(b)} Linear scaling of the unit-circle root
  count $N_{\mathrm{uc}}(P_d) \approx 0.8251\,d$ with polynomial degree $d$
  across two orders of magnitude.}
  \label{fig:wilkinson_sturm}
\end{figure}

\begin{figure}[t!]
  \centering
  \includegraphics[width=0.96\textwidth]{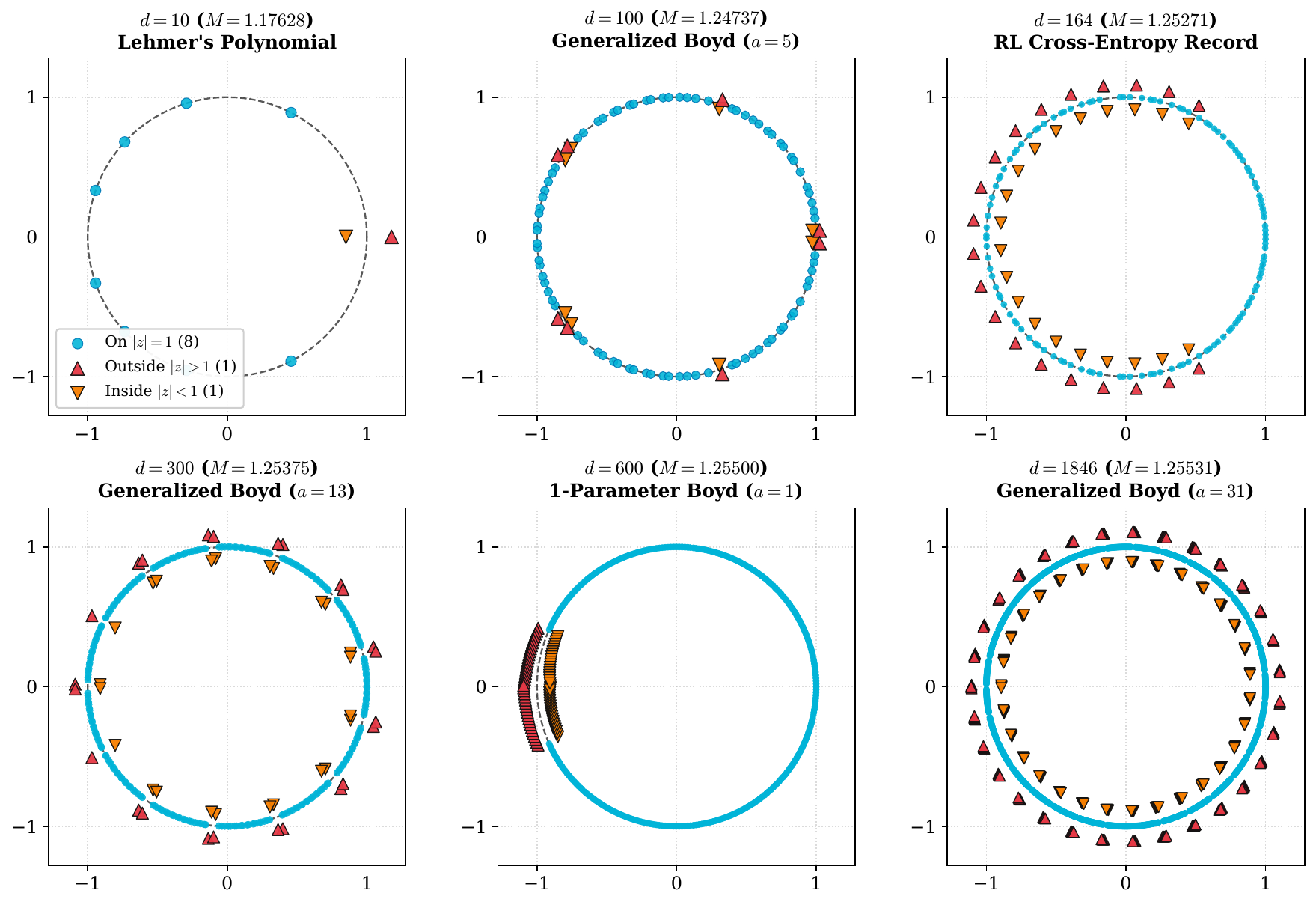}
  \caption{\textbf{Complex-plane root distributions for representative
  polynomials ($d \in \{10, 100, 164, 300, 600, 1846\}$).} Most roots lie on
  the unit circle $|z|=1$ (cyan circles), while the remaining roots form
  reciprocal pairs outside ($|z|>1$, red triangles) and inside ($|z|<1$, orange
  triangles) the circle. At $d=10$ (Lehmer's polynomial), only a single pair
  lies off the circle; at higher degrees, off-circle roots either concentrate
  near $z=-1$ ($d=164$ and $d=600$, $a=1$) or split into $a$ evenly spaced
  clusters around the circle ($d=100, 300, 1846$ with $a=5, 13, 31$). Radial
  distances from $|z|=1$ are magnified for visibility.}
  \label{fig:constellations}
\end{figure}

Now that we have a general algorithm for constructing high degree polynomials 
with minimal Mahler measure, it is natural to ask
why these polynomials achieve Mahler measures so close
to $1$, and what physical mechanism governs the arrangement of their roots in
the complex plane $\mathbb{C}$ as $d \to \infty$?

To get a physical interpretation, we first use Jensen's formula to rewrite the
logarithmic Mahler measure~\eqref{eq:mahler_definition} of a monic polynomial
$P(x) = \prod_{j=1}^d (x - z_j)$ as a unit-circle integral\footnote{By Jensen's
formula, $\frac{1}{2\pi}\int_0^{2\pi} \log|P(e^{i\theta})|\,d\theta =
\log|P(0)| - \sum_{|z_j| < 1} \log|z_j|$. Since $\log|P(0)| = \sum_{j=1}^d
\log|z_j|$ is the sum over all roots, subtracting the roots inside the unit
circle $|z_j| < 1$ leaves $\sum_{|z_j| > 1} \log|z_j|$ (as $\log|z_j| = 0$ on
$|z_j| = 1$).}:
\begin{equation}
  \log M(P) = \sum_{|z_j| > 1} \log |z_j| = \frac{1}{2\pi} \int_0^{2\pi} \log
  |P(e^{i\theta})| \, d\theta = \frac{1}{2\pi} \int_0^{2\pi} \sum_{j=1}^d \log
  \big| e^{i\theta} - z_j \big| \, d\theta.
  \label{eq:jensen_mahler_integral}
\end{equation}
Next, we recall that in two dimensions, Gauss's law over a circle of
circumference $2\pi r$ gives an electric field $E \propto 1/r$, so integrating
$\int dr/r$ gives a logarithmic potential $\Phi(x) = -\sum_{j=1}^d q_j
\log|x - z_j|$. For unit charges ($q_j = 1$) sitting at the roots $\{z_1,
\dots, z_d\}$, the electric potential at any probe point $x$ is $\Phi(x) =
- \sum_{j=1}^d \log|x - z_j| = - \log|P(x)|$. Averaging this potential around
the unit circle $x = e^{i\theta}$ recovers the log Mahler measure
$\log M(P) = \sum_{|z_j|>1} \log|z_j|$. This gives a clear physical
interpretation: \emph{minimizing the Mahler measure over irreducible
polynomials of degree $d$ is equivalent to finding the low-energy equilibrium
of $d$ unit charges subject to the discrete integer coefficient lattice
$\mathbb{Z}^{d+1}$}.

Without an external confining force, $d$ points with the same charges
($q_j = +1$) repel each another out to infinity.
In our case, minimizing $\log M(P)$ over
irreducible reciprocal polynomials in $\mathbb{Z}[x]$ creates two radial walls
that trap the charges around the unit circle $|z| = 1$:
\begin{itemize}[leftmargin=*,nosep]
  \item \textbf{Outer wall ($|z_j| > 1$):} Minimizing $\log M(P) =
    \sum_{j=1}^d \log\max(1, |z_j|)$ imposes a radial energy penalty
    $\log|z_j| > 0$ on every exterior root, pushing roots outside the unit disk
    inward toward $|z| = 1$ so they cannot escape to infinity.
  \item \textbf{Inner wall ($|z_j| < 1$):} Reciprocal symmetry $z_j
    \leftrightarrow z_j^{-1}$ pairs every interior root at radius $r_j < 1$
    with an exterior partner at $r_j^{-1} > 1$ carrying penalty $\log(1/r_j) >
    0$, pushing interior roots outward so they cannot collapse to the origin.
\end{itemize}
Finally, once the charges are trapped near the circular wire $|z| = 1$,
irreducibility and integer coefficients supply the angular repulsion that
spreads them apart. Without irreducibility, all $d$ charges could collapse onto
a single point (such as $(x-1)^d$). Because $P \in \mathbb{Z}[x]$ is
irreducible, its roots are strictly distinct, and integer coefficients force
the product of their squared distances $\prod_{j<k}|z_j - z_k|^2$ to be a
nonzero integer ($\ge 1$); this pairwise repulsion prevents clustering and
drives uniform angular spacing on $|z| = 1$ as $d \to
\infty$~\cite{bilu1997equidistribution}. At finite degree $d$, this balance
between radial confinement and angular repulsion packs over $80\%$ of the $d$
roots onto $|z| = 1$, with mean angular spacing $\langle \Delta\theta \rangle
\approx 2\pi / (0.825\,d)$.

Figure~\ref{fig:wilkinson_sturm} confirms this picture across all 824 benchmark
polynomials: a nearly constant fraction of roots---between $77.0\%$ and
$88.0\%$, with a mean of $82.51\%$---lie \emph{exactly} on the unit circle
$|z| = 1$ (verified via exact rational root counting in
Appendix~\ref{app:unit_circle_sturm}, Table~\ref{tab:sturm_audit}). The
remaining $\approx 17.49\%$ of roots sit just off $|z| = 1$ in reciprocal pairs
at distance $\log|z_j| = \mathcal{O}(1/d)$, forming either a single localized
cluster ($a = 1$) or $a$ symmetric clusters ($a > 1$) around the circle, as
illustrated across representative degrees in Figure~\ref{fig:constellations}
(see Appendix~\ref{app:off_circle_clusters} and Figure~\ref{fig:off_circle} for
further details).
\section{Benchmark Dataset of Minimal-Mahler Polynomials}
\label{sec:dataset}

To make our computational and analytic findings directly accessible to the
mathematical community, we assemble and release a verified benchmark dataset
(\texttt{mahler\_\allowbreak dataset\_\allowbreak d10\_\allowbreak 2000.csv})
covering all 996 even degrees $d \in [10, 2000]$. This dataset extends the previous   work~\cite{lehmer1933factorization,boyd1981speculations,boyd1989reciprocal,mossinghoff1998list,mossinghoff2008minimal}, which systematically reached $d \le 180$ (with isolated records up to $d \le 250$).

\paragraph{Polynomial Representations.}
A practical challenge in tabulating high-degree irreducible polynomials is
representation size: while a sparse polynomial $Q_{a,b,k,s_0,s_1}(x)$ in Boyd's
family contains only 7 non-zero terms, dividing out its cyclotomic factors
$\Phi(x)$ produces a dense irreducible quotient polynomial $P_d^*(x) =
Q_{a,b,k,s_0,s_1}(x)/\Phi(x) \in \mathbb{Z}[x]$ with up to $\sim 1{,}000$
non-zero coefficients at $d = 2000$. To support direct computer-algebra use as
well as compact inspection, entries in the dataset are recorded in three
complementary forms:
\begin{enumerate}[leftmargin=*,nosep]
  \item \textbf{Expanded polynomial string (\texttt{polynomial}):} Every degree
    entry provides the dense reciprocal polynomial string $P_d^*(x) =
    \sum_{j=0}^d a_j x^j \in \mathbb{Z}[x]$ written out explicitly in powers of
    $x$ (e.g., \texttt{x\textasciicircum 10 + x\textasciicircum 9 -
    x\textasciicircum 7 - \dots\ + 1}).
  \item \textbf{Explicit sparse quotient expression (\texttt{quotient\_expr}):}
    For the $99.6\%$ of degrees in $d \in [12, 2000]$ (991 of 995 degrees)
    where the Generalized Boyd family attains or ties the best-known record, we
    record the self-contained algebraic fraction
    $Q_{a,b,k,s_0,s_1}(x)/\Phi(x)$ explicitly
    in powers of $x$. In Mathematica, wrapping any entry in
    \texttt{Cancel[\dots]} performs exact polynomial division over
    $\mathbb{Z}[x]$ to recover the expanded polynomial $P_d^*(x)$, while the
    7-term numerator alone determines $M(P_d^*)$ since
    $M(\Phi) = 1$.\footnote{For instance, at $d = 1000$, evaluating
    \texttt{Cancel[(x\textasciicircum 1008 - x\textasciicircum 991 -
    x\textasciicircum 521 + x\textasciicircum 504 - x\textasciicircum 487 -
    x\textasciicircum 17 + 1) / (x\textasciicircum 8 - x\textasciicircum 7 +
    x\textasciicircum 5 - x\textasciicircum 4 + x\textasciicircum 3 - x + 1)]}
    (or dividing by \texttt{Cyclotomic[15, x]}) directly yields the 537-term
    irreducible polynomial $P_{1000}^*(x)$, and solving \texttt{NSolve} on the
    7-term numerator gives $M(P_{1000}^*) \approx 1.25513074$ without
    expanding.}
  \item \textbf{Five-parameter tuple and cyclotomic divisor
    (\texttt{boyd\_tuple}, \texttt{cyclotomic\_divisor}):} For every
    Generalized Boyd entry, we also record the five integer parameters
    $(a, b, k, s_0, s_1)$ (with $k = 0$ and $a \le b$ across all 991 records)
    and factored cyclotomic divisor $\Phi(x) = \prod_{\Phi_m \mid
    Q_{a,b,k,s_0,s_1}} \Phi_m(x)$ of degree $\delta = 2b + 2a + k - d$ such
    that
    \begin{equation}
      P_d^*(x) = \frac{x^{a+k}(x^a + s_0)x^{2b} + s_1(x^{2a} + s_0 x^a +
      1)x^b + s_0(x^a + s_0)}{\Phi(x)}.
      \label{eq:dataset_quotient_form}
    \end{equation}
\end{enumerate}

\paragraph{Verification and Provenance.}
Every entry in the dataset for $d \in [10, 2000]$ was selected by taking the
minimum Mahler measure across our search methods, the one-parameter Boyd
family, and the Generalized Boyd algorithm (together with published records for
$d \le 250$), and verified using the protocol in
Section~\ref{sec:efficient-search}. As summarized in
Table~\ref{tab:boyd_improvements}, the Generalized Boyd family accounts for 991
of the 996 records ($99.6\%$ of $d \in [12, 2000]$, including all $d \ge 166$),
while our search methods provide the remaining five optima ($d \in \{10, 12,
16, 32, 164\}$).

\paragraph{Notable Benchmark Records.}

Table~\ref{tab:dataset_highlights} presents a subset of notable records from
the dataset spanning three degree regimes across $10 \le d \le 2000$. At small
degrees ($10 \le d \le 28$), the dataset is anchored at $d = 10$ and $d = 12$
by Lehmer's degree-10 polynomial $L(x)$ and the degree-12 evolutionary optimum,
while the five-parameter Generalized Boyd family already captures the global
optimum at $d = 14$ and $d = 18$. In the intermediate regime ($30 \le d \le
250$), within the range covered by published in the literature, our dataset
matches 85 known optima and establishes strict improvements on 6 benchmark
degrees: $d = 178$ (inside the systematically searched $d \le 180$ regime) and
$d = 188$, $190$, $202$, $242$, and $250$ (see Table~\ref{tab:lit_improvements}).
Finally, all 996 optimal polynomials in $d \in [10, 2000]$ achieve a Mahler measure
strictly below Boyd's infinite-degree limit $\alpha_1 \approx 1.25543387$,
approaching $\alpha_1$ asymptotically from below as $d$ increases.

\begin{table}[t!]
  \centering
  \caption{\textbf{Selected Benchmark Records from the 996-Degree Dataset ($d
  \in [10, 2000]$)}, illustrating small-degree optima, all 6 strict
  improvements over published Literature Optima, and high-degree records in
  compact cyclotomic-quotient form.}
  \label{tab:dataset_highlights}
  \small
  \begin{tabularx}{\textwidth}{@{}r r r l >{\raggedright\arraybackslash}X r@{}}
    \toprule
    \textbf{Deg $d$} & \textbf{Best $M(P_d^*)$} &
    \textbf{Lit.\ $M_{\mathrm{lit}}$} & \textbf{Method} &
    \textbf{Tuple $(a,b,k,s_0,s_1) / \Phi$ or Form} &
    \textbf{$N_{\mathrm{uc}}(P_d^*)/d$} \\
    \midrule
    $10$ & $1.17628082$ & $1.17628082$ & Tiling / Lit &
    Lehmer's polynomial $L(x)$~\eqref{eq:lehmer_poly} & $80.0\%$ \\
    $12$ & $1.22778556$ & $1.22778556$ & Evol.\ / Lit &
    $x^{12}+x^{11}+x^{10}-x^8-x^7-x^6-x^5-x^4+x^2+x+1$ & $66.7\%$ \\
    $14$ & $1.20002652$ & $1.20002652$ & Gen.\ Boyd &
    $(3, 4, 0, -1, -1)$ (7-term, $\delta=0$) & $85.7\%$ \\
    $18$ & $1.18836815$ & $1.18836815$ & Gen.\ Boyd / RL &
    $(1, 10, 0, -1, -1) / \Phi_{12}$ ($\delta=4$) & $88.9\%$ \\
    \midrule
    $178$ & $1.25162839$ & $1.25204881$ & Gen.\ Boyd &
    $(4, 89, 0, -1, -1) / \Phi_{15}$ ($\delta=8, \Delta M = -0.00042$) &
    $85.4\%$ \\
    $188$ & $1.25207822$ & $1.28309200$ & Gen.\ Boyd / RL &
    $(11, 86, 0, +1, +1) / \Phi_7$ ($\delta=6, \Delta M = -0.03101$) &
    $83.0\%$ \\
    $190$ & $1.25198376$ & $1.28518461$ & Gen.\ Boyd &
    $(2, 97, 0, -1, +1) / \Phi_{30}$ ($\delta=8, \Delta M = -0.03320$) &
    $86.3\%$ \\
    $202$ & $1.25253255$ & $1.25478722$ & Gen.\ Boyd &
    $(11, 93, 0, -1, -1) / \Phi_{14}$ ($\delta=6, \Delta M = -0.00225$) &
    $83.2\%$ \\
    $230$ & $1.25292932$ & --- & Gen.\ Boyd &
    $(24, 91, 0, +1, -1)$ (7-term, $\delta=0$) & $80.9\%$ \\
    $242$ & $1.25305628$ & $1.28571755$ & Gen.\ Boyd &
    $(10, 111, 0, -1, -1)$ ($\delta=0, \Delta M = -0.03266$) & $84.3\%$ \\
    $250$ & $1.25316520$ & $1.28497453$ & Gen.\ Boyd &
    $(2, 127, 0, -1, +1) / \Phi_{30}$ ($\delta=8, \Delta M = -0.03181$) &
    $86.4\%$ \\
    \midrule
    $346$ & $1.25392132$ & --- & Gen.\ Boyd &
    $(10, 163, 0, +1, +1)$ (7-term, $\delta=0$) & $85.0\%$ \\
    $442$ & $1.25449510$ & --- & Gen.\ Boyd &
    $(14, 207, 0, +1, +1)$ (7-term, $\delta=0$) & $84.6\%$ \\
    $526$ & $1.25466438$ & --- & Gen.\ Boyd &
    $(11, 256, 0, +1, -1) / \Phi_{30}$ ($\delta=8$) & $85.2\%$ \\
    $746$ & $1.25498751$ & --- & Gen.\ Boyd &
    $(1, 372, 0, +1, -1)$ (7-term, $\delta=0$) & $86.9\%$ \\
    $1000$ & $1.25513074$ & --- & Gen.\ Boyd &
    $(17, 487, 0, -1, -1) / \Phi_{15}$ ($\delta=8$) & $85.4\%$ \\
    $1500$ & $1.25528161$ & --- & Gen.\ Boyd &
    $(139, 611, 0, -1, -1)$ (7-term, $\delta=0$) & $80.7\%$ \\
    $2000$ & $1.25533130$ & --- & Gen.\ Boyd &
    $(77, 923, 0, -1, +1)$ (7-term, $\delta=0$) & $83.8\%$ \\
    \bottomrule
  \end{tabularx}
\end{table}

\section{Discussion}
\label{sec:discussion}

Our results show that specializations of Boyd's two-variable polynomial
$F(X, Y)$ yield verified irreducible polynomials with Mahler measure strictly
below $\alpha_1 \approx 1.255434$ across all tested even degrees $d \in [12,
2000]$ and follow the asymptotic lower envelope $\alpha_1 - A d^{-3/2}$ as $d
\to \infty$. This explains the common asymptote reached by independent numerical
searches. Nevertheless, key
questions around Lehmer's problem remain open. In particular, does $\alpha_1$
represent the true limit infimum of $M(P)$ as $d \to \infty$, or can isolated
polynomials with Mahler measure substantially below $\alpha_1$---such as
Lehmer's polynomial at $d = 10$ ($M \approx 1.17628$)---still occur at
arbitrarily large degrees? A natural next step is to test whether similar
multi-parameter cyclotomic constructions can systematically generate small
Salem numbers (where only a single root lies outside $|z| = 1$) or extend to
other two- and multi-variable Mahler limit
points~\cite{boyd1981speculations,boyd1998mahler}.

More broadly, this work illustrates how a human--AI discovery loop can move
beyond black-box search that stops at a numerical bound or an isolated example.
Here, large-scale numerical optimization served as an empirical microscope:
finding hundreds of high-degree polynomials revealed repeating coefficient
patterns that human and AI researchers could then compress via cyclotomic
factoring into an exact analytical construction. Combining broad computational
search, automated verification, and analytical synthesis offers a practical
template across number theory, combinatorics, theoretical physics, biology, and
other domains where large-scale computation can uncover hidden structure behind
complex systems.

\appendix

\section{Asymptotic Analysis of the Generalized Boyd Family}
\label{app:stationary_phase}

\subsection{Convergence to Boyd's Limit Point}
\label{app:boyd_limit}

As observed in Section \ref{sec:search_algorithms}, the best high-degree
polynomials across five different search methods approach the asymptote $\alpha_1
\approx 1.255434$ and collapse to Generalized Boyd quotients $P_d(x) = F(s_0
x^a, s_1 x^b)/\Phi(x)$. In this section we show
analytically why $M(P_d)$ converges to $\alpha_1 = M(F(X,Y))$ as $d \to
\infty$. Here as before, $F(X, Y) = X(X+1)Y^2 + (X^2+X+1)Y + (X+1)$ is Boyd's
two-variable polynomial introduced in~\cite{boyd1981speculations} and defined
in~\eqref{eq:boyd_f_xy}.

First, we note that substituting $(X, Y) = (s_0 x^a, s_1 x^b)$ into the
leading term $X^2 Y^2$ of $F(X, Y)$ gives degree $2a + 2b$, so the Generalized
Boyd polynomial has a degree $d = 2a + 2b - \deg\Phi \approx 2a + 2b$.
Since the Generalized Boyd algorithm uses small $a$ (such as $a \le 20$ in
Section~\ref{sec:analytic_resolution}), growing $d \to \infty$ sends $b/a
\approx d/(2a) \to \infty$.
Secondly, because cyclotomic
polynomials have Mahler measure $M(\Phi) = 1$, multiplicativity ($M(P_d \cdot
\Phi) = M(P_d)M(\Phi) = M(P_d)$) implies that $M(P_d)$ equals the Mahler
measure of the one-variable specialization $F(s_0 x^a, s_1 x^b)$.

Next, by Jensen's formula~\eqref{eq:jensen_mahler_integral}, $\log M(F(s_0 x^a,
s_1 x^b))$ averages $\log|F(X, Y)|$ along the one-dimensional path $(X, Y) =
(s_0 e^{ia\theta}, s_1 e^{ib\theta})$, whereas the two-variable Mahler measure
$M(F(X, Y))$ averages $\log|F|$ over the entire unit two-torus $\mathbb{T}^2 =
\{(e^{it}, e^{i\tau}) : t, \tau \in [0, 2\pi)\}$:
\begin{equation}
  \begin{aligned}
    \log M\bigl(F(s_0 x^a, s_1 x^b)\bigr) &= \frac{1}{2\pi} \int_0^{2\pi}
    \log\bigl| F(s_0 e^{ia\theta}, s_1 e^{ib\theta}) \bigr| \, d\theta, \\
    \log M(F(X, Y)) &= \frac{1}{(2\pi)^2} \int_0^{2\pi} \int_0^{2\pi}
    \log\bigl| F(e^{it}, e^{i\tau}) \bigr| \, dt \, d\tau.
  \end{aligned}
  \label{eq:two_var_mahler_def}
\end{equation}
Then, by the limit theorems of Boyd~\cite{boyd1981speculations} and
Lawton~\cite{lawton1983problem}, as $b/a \to \infty$, the one-dimensional path
equidistributes on $\mathbb{T}^2$, so the path integral converges to the torus
integral:
\begin{equation}
  \lim_{b/a \to \infty} M\big( F(s_0 x^a, s_1 x^b) \big) = M(F(X, Y)) =
  \alpha_1 = 1.255433866266608745189\dots
  \label{eq:exact_c_value}
\end{equation}
This value $\alpha_1 = M(F(X, Y))$ is the smallest known limit point of Mahler
measures~\cite{boyd1981speculations,boyd2005small}.

Finally, we note that geometrically, the zero locus $F(X, Y) = 0$ defines an
elliptic curve $E_{15}$ of conductor $15$, whose logarithmic Mahler measure can
be evaluated to arbitrary precision via its Hasse--Weil $L$-function at $s =
2$~\cite{deninger1997deligne,boyd1998mahler}:
\begin{equation}
  \log \alpha_1 = \frac{5\sqrt{15}}{4\pi^2} L(E_{15}, 2) =
  0.227481223003190\dots
\end{equation}

\subsection{Universal Bound on Cyclotomic Factors ($\delta \le 14$)}
\label{app:cyclotomic_bound}

In Section~\ref{sec:analytic_resolution}, we stated that sparse seven-term
polynomials $Q_{a,b,k,s_0,s_1}(x)$ with $\gcd(a, b, k) = 1$ only admit
low-degree cyclotomic factors $\Phi(x)$ with $\delta = \deg\Phi(x) \le 14$,
arising from the five divisors $\Phi_7, \Phi_{12}, \Phi_{14}, \Phi_{15},
\Phi_{30}$. Recall from Smyth's bound~\eqref{eq:smyth_bound} that any
irreducible polynomial $P(x) = Q_{a,b,k,s_0,s_1}(x)/\Phi(x)$ with $M(P) <
1.324718$ must be reciprocal. Next, since every cyclotomic factor $\Phi(x)$ is
reciprocal up to sign, $Q_{a,b,k,s_0,s_1}(x)$ must also be reciprocal, which
holds if and only if $k = 0$. Here we prove that for the reciprocal family
$k = 0$, the bound $\delta \le 14$ is an exact, unconditional property for all
coprime integers $a, b \ge 1$ and arbitrarily large degrees $d$.

At $k = 0$, we have $Q_{a,b,0,s_0,s_1}(x) = F(s_0 x^a, s_1 x^b)$, where
$F(X, Y) = X(X+1)Y^2 + (X^2+X+1)Y + (X+1)$ is symmetric in $X$ and $Y$
($F(X, Y) = F(Y, X)$). Suppose $\Phi_m(x)$ divides $F(s_0 x^a, s_1 x^b)$ with
$\gcd(a, b) = 1$, and let $\zeta_m = e^{2\pi i/m}$ be a primitive $m$-th root
of unity. Setting $X = s_0 \zeta_m^a$ and $Y = s_1 \zeta_m^b$, the pair
$(X, Y)$ is a torsion point on the algebraic curve $F(X, Y) = 0$, where $X$ and
$Y$ are primitive $m_X$-th and $m_Y$-th roots of unity ($X^{m_X} = Y^{m_Y} =
1$). Since $\gcd(a, b) = 1$, B\'ezout's identity $u a + v b = 1$ ($u, v \in
\mathbb{Z}$) gives
\begin{equation}
  \zeta_m = (\zeta_m^a)^u (\zeta_m^b)^v = (s_0 X)^u (s_1 Y)^v.
  \label{eq:bezout_zeta}
\end{equation}
Because $s_0^2 = s_1^2 = X^{m_X} = Y^{m_Y} = 1$, raising both sides to the
least common multiple $L = \operatorname{lcm}(2, m_X, m_Y)$ yields $\zeta_m^L =
1$, so $m$ must divide $\operatorname{lcm}(2, m_X, m_Y)$. While the finiteness
of torsion points on an irreducible genus-$1$ curve $F(X, Y) = 0$ in
$\mathbb{G}_m^2$ follows from Lang's theorem~\cite{bombieri2006heights}, we can
determine all allowed orders $(m_X, m_Y)$ and their exact combinations in four steps:
\begin{enumerate}
  \item \textbf{Algebraic unit condition.} Writing $X^2 + X + 1 = X(X+1) + 1$
    allows us to rearrange $F(X, Y) = 0$ as
    \begin{equation}
      (X + 1)(X Y^2 + X Y + 1) + Y = 0 \implies \frac{1}{X + 1} = -(X Y + X +
      Y^{-1}).
    \end{equation}
    Because $X$ and $Y$ are roots of unity, $X, Y, Y^{-1}$ are algebraic
    integers; hence $(X + 1)^{-1}$ is an algebraic integer, so $X + 1$ (and by
    symmetry $Y + 1$) must be an algebraic unit. For a primitive $m_X$-th root
    of unity $X$, $\operatorname{Norm}_{\mathbb{Q}(X)/\mathbb{Q}}(X + 1) =
    \Phi_{m_X}(-1)$ is a unit ($\pm 1$) if and only if $m_X \ne 1$, $m_X \ne
    2^r$ ($r \ge 1$), and $m_X \ne 2 p^s$ for any odd prime $p$ and $s \ge 1$.
  \item \textbf{Galois conjugate discriminant bound.} For any Galois
    automorphism $\sigma$ of $\mathbb{Q}(\zeta_m)/\mathbb{Q}$,
    the conjugate $(\sigma(X), \sigma(Y))$ also satisfies $F(\sigma(X),
    \sigma(Y)) = 0$ with $|\sigma(X)| = |\sigma(Y)| = 1$. Rearranging $F = 0$
    as $(X^2 + X + 1)Y = -X(X+1)Y^2 - (X+1)$ and applying the triangle
    inequality on $|X| = |Y| = 1$ gives $|X^2 + X + 1| \le 2|X + 1|$. Setting
    $\sigma(X) = e^{it}$ with $t = 2\pi r/m_X$ ($\gcd(r, m_X) = 1$) and
    squaring $|2\cos t + 1| \le 4|\cos(t/2)|$ yields $4\cos^2 t - 4\cos t - 7
    \le 0$ (see also
    Eqs.~\eqref{eq:rescaled_boyd_quadratic}--\eqref{eq:boyd_branch_angle}), i.e.,
    \begin{equation}
      \cos\!\left(\frac{2\pi r}{m_X}\right) \ge \frac{1}{2} - \sqrt{2} \approx
      -0.91421 \qquad \text{for all } \gcd(r, m_X) = 1.
    \end{equation}
    Choosing the coprime residue $r$ closest to $m_X/2$ immediately bounds
    $m_X$:
    \begin{itemize}
      \item For odd $m_X$: $r = (m_X - 1)/2$ is coprime to $m_X$, giving
        $-\cos(\pi/m_X) \ge \frac{1}{2} - \sqrt{2} \implies m_X \le 7$. Step 1
        eliminates $m_X = 1$, leaving $m_X \in \{3, 5, 7\}$.
      \item For $m_X \equiv 0 \pmod 4$: $r = m_X/2 - 1$ is coprime to $m_X$,
        giving $-\cos(2\pi/m_X) \ge \frac{1}{2} - \sqrt{2} \implies m_X \le 12$.
        Step 1 eliminates $4$ and $8$, leaving $m_X = 12$.
      \item For $m_X \equiv 2 \pmod 4$: $r = m_X/2 - 2$ is coprime to $m_X$,
        giving $-\cos(4\pi/m_X) \ge \frac{1}{2} - \sqrt{2} \implies m_X \le 30$.
        Step 1 eliminates all $2p^s \in \{2, 6, 10, 14, 18, 22, 26\}$, leaving
        only $m_X = 30$.
    \end{itemize}
    By symmetry $F(X, Y) = F(Y, X)$, both $m_X$ and $m_Y$ must belong to
    $\{3, 5, 7, 12, 30\}$.
  \item \textbf{Exact classification of cyclotomic divisors.} Solving the
    quadratic $F(X, Y) = 0$ at $m_X \in \{3, 5, 7, 12, 30\}$ yields three
    torsion families, governed by three exact polynomial identities (together
    with $X \leftrightarrow Y$ symmetry and Vieta inversion $F(X, Y) = X^3 Y^2
    F(X, X^{-1}Y^{-1})$):
    \begin{equation}
      \begin{aligned}
        F(X, X^2) &= X^5 F(X, X^{-3}) = \Phi_7(X) \quad
          (m_X = m_Y = 7), \\
        F(X, -X) &= X^3 F(X, -X^{-2}) = \Phi_{12}(X) \quad
          (m_X, m_Y \in \{3, 12\}), \\
        F(X, -X^3) &= X^7 F(X, -X^{-4}) = \Phi_{30}(X) \quad
          (m_X, m_Y \in \{5, 30\}),
      \end{aligned}
      \label{eq:three_torsion_families}
    \end{equation}
    where $(m_X, m_Y) = (3, 12)$ and $(5, 30)$ arise from the $X \leftrightarrow
    Y$ transposes $X = -Y^{-2}$ and $X = -Y^3$. By~\eqref{eq:bezout_zeta}, the
    first family ($m_X = m_Y = 7$) yields only $\Phi_7$ or $\Phi_{14}$ (degree
    $6$); the second family ($\operatorname{lcm}(m_X, m_Y) = 12$) yields only
    $\Phi_{12}$ (degree $4$); and the third family ($\operatorname{lcm}(m_X,
    m_Y) = 30$) yields only $\Phi_{15}$ or $\Phi_{30}$ (degree $8$). Moreover, a
    repeated factor $\Phi_m(x)^2 \mid F(s_0 x^a, s_1 x^b)$ would require $a X
    F_X(X, Y) + b Y F_Y(X, Y) = 0$ across every Galois conjugate $(\sigma(X),
    \sigma(Y))$, forcing the logarithmic slope $-X F_X / (Y F_Y) = b/a \in
    \mathbb{Q}_{>0}$ to be rational and positive; however, evaluating at the
    primitive parameter root $e^{2\pi i/k}$ ($k \in \{7, 12, 30\}$) on every
    branch of~\eqref{eq:three_torsion_families} gives $-X F_X / (Y F_Y) < 0$,
    so every cyclotomic factor has multiplicity $1$.
  \item \textbf{Mutual exclusivity and sharp bound $\delta \le 14$.} For
    $\gcd(a, b) = 1$, $a$ and $b$ cannot both be even, so $x \mapsto -x$ flips
    at least one sign in $(s_0, s_1) \mapsto (s_0(-1)^a, s_1(-1)^b)$.
    Consequently:
    \begin{itemize}
      \item $\Phi_7$ (which requires $(s_0, s_1) = (+1, +1)$) and
        $\Phi_{14}(x) = \Phi_7(-x)$ cannot both divide $Q$, contributing at
        most degree $6$ (when $b \equiv 2a$ or $4a \pmod 7$).
      \item Similarly, $\Phi_{15}$ and $\Phi_{30}(x) = \Phi_{15}(-x)$ cannot
        both divide $Q$, contributing at most degree $8$.
      \item Finally, $\Phi_{12}$ and $\{\Phi_{15}, \Phi_{30}\}$ can never
        simultaneously divide $Q$ because their congruences modulo $3$ are
        disjoint: every branch of $\Phi_{12}$ in~\eqref{eq:three_torsion_families}
        ($Y = -X$, $Y = -X^{-2}$, $X = -Y^{-2}$) requires $3 \nmid a$, $3 \nmid
        b$, and $b \equiv a \pmod 3$, whereas the branches of $\Phi_{15}$ and
        $\Phi_{30}$ require either $3 \mid ab$ (for $Y = -X^3$ or $X = -Y^3$,
        where one coordinate has order $5$) or $b \equiv -4a \equiv 2a \pmod 3$
        (for $Y = -X^{-4}$).
    \end{itemize}
\end{enumerate}
Combining these bounds, $\delta = \deg\Phi(x)$ can only take values in
$\{0, 4, 6, 8, 10, 14\}$, with global maximum $\delta \le 6 + 8 = 14$ across
all coprime $(a, b)$ and all degrees $d$.

\subsection{Existence of Self-Consistent Solutions for Every Even Degree}
\label{app:self_consistent_existence}

Having classified all possible cyclotomic factors of $Q_{a,b,0,s_0,s_1}(x)$ in
Section~\ref{app:cyclotomic_bound}, we now address a natural question raised by
the Generalized Boyd algorithm in Section~\ref{sec:analytic_resolution}: why
does solving~\eqref{eq:exact_diophantine_b} for $b = (d + \delta - 2a -
k)/2$ guarantee at least one valid, self-consistent solution at every target
even degree $d$? Because the cyclotomic degree $\delta = \deg\Phi(x)$ is itself
determined by $(a, b, k, s_0, s_1)$, a candidate tuple with $b = (d + \delta -
2a - k)/2$ produces a quotient $P(x) = Q_{a,b,k,s_0,s_1}(x)/\Phi(x)$ of exact
degree $d$ if and only if the actual cyclotomic factor $\Phi(x)$ of
$Q_{a,b,k,s_0,s_1}(x)$ has degree $\deg\Phi(x) = \delta$. Using the divisibility
conditions from Section~\ref{app:cyclotomic_bound}, we show below that such a
self-consistent tuple is guaranteed to exist for every even degree $d = 2m \ge
12$ ($m \ge 6$).

First, observe that for any fixed coprime pair $(a, b)$ with $b \not\equiv 2a,
4a \pmod 7$, at least two of the four sign pairs $(s_0, s_1) \in \{-1, +1\}^2$
have $\delta = 0$. Indeed, when $b \not\equiv 2a, 4a \pmod 7$, neither $\Phi_7$
nor $\Phi_{14}$ can divide $F(s_0 x^a, s_1 x^b)$ for any $(s_0, s_1)$, and by
Step 4 of Section~\ref{app:cyclotomic_bound}, $\Phi_{12}$ and $\{\Phi_{15},
\Phi_{30}\}$ cannot both occur for the same $(a, b)$. Moreover, whichever
family occurs hits at most two sign pairs:
\begin{itemize}
  \item For $\{\Phi_{15}, \Phi_{30}\}$, the three congruences $b \equiv 3a$, $a
    \equiv 3b$, and $b \equiv -4a \pmod{15}$ are mutually exclusive for
    $\gcd(a, b) = 1$; the single active congruence uniquely fixes $(s_0, s_1)$
    at $x = \zeta_{15}$ and its $x \mapsto -x$ partner at $x = \zeta_{30}$ (at
    most two sign pairs).
  \item For $\Phi_{12}$ (where $3 \nmid ab$ and $b \equiv a \pmod 3$), setting
    $x = \zeta_{12}$ ($\zeta_{12}^6 = -1$) reduces the three branches to
    $\zeta_{12}^{b-a} = -s_0 s_1$, $\zeta_{12}^{2a+b} = -s_1$, or
    $\zeta_{12}^{a+2b} = -s_0$. Modulo $6$, either $b - a \equiv 0 \pmod 6$
    (activating only $s_0 s_1 = -1$, since $2a + b \equiv a + 2b \equiv 3 \pmod
    6$) or $b - a \equiv 3 \pmod 6$ (where $(2a+b) + (a+2b) \equiv 3 \pmod 6$,
    so exactly one of $2a + b$ or $a + 2b$ is $0 \pmod 6$, fixing either $s_1$
    or $s_0$ and hitting two sign pairs).
\end{itemize}
By the Pigeonhole Principle, at least two of the four sign pairs $(s_0, s_1)
\in \{-1, +1\}^2$ must be free of all cyclotomic factors ($\delta = 0$).

Now, for any integer $m = d/2 \ge 6$, set $k = 0$, $\delta = 0$, and choose
\begin{equation}
  a = \begin{cases}
    1, & \text{if } m \not\equiv 3, 5 \pmod 7, \\
    7, & \text{if } m \equiv 3 \text{ or } 5 \pmod 7,
  \end{cases}
  \qquad b = m - a.
  \label{eq:explicit_self_consistent_a}
\end{equation}
When $m \not\equiv 3, 5 \pmod 7$, $a = 1$ gives $\gcd(a, b) = \gcd(1, m - 1) =
1$, $b \ge 5$, and $b = m - 1 \not\equiv 2, 4 \pmod 7$. When $m \equiv 3$ or
$5 \pmod 7$ (which forces $m \ge 10$ since $m \ge 6$), we have $7 \nmid m$, so
$a = 7$ satisfies $\gcd(a, b) = \gcd(7, m) = 1$, $b = m - 7 \ge 3$, and $b
\equiv m \equiv 3, 5 \not\equiv 0 \equiv 2a, 4a \pmod 7$ (with $Q_{7,3,0,s_0,s_1}
= Q_{3,7,0,s_1,s_0}$ at $m = 10$ by $X \leftrightarrow Y$ symmetry if one
orders $a \le b$). Thus, in both cases,
$(a, b)$ is a valid coprime pair with $b \not\equiv 2a, 4a \pmod 7$,
guaranteeing at least two sign pairs $(s_0, s_1) \in \{-1, +1\}^2$ for which
$Q_{a,b,0,s_0,s_1}(x)$ has $\delta = 0$ and exact degree $d = 2a + 2b$.
Sweeping $\delta \in \{0, 4, 6, 8, 10, 14\}$ and larger $a$ in~\eqref{eq:exact_diophantine_b} is therefore not required for existence, but
rather enlarges the candidate pool to include cyclotomic quotients ($\delta >
0$) when they achieve a smaller Mahler measure at degree $d$.

\subsection{Oscillatory $d^{-3/2}$ Asymptotics}
\label{app:trough_scaling}

In Section~\ref{app:boyd_limit}, we saw that $\log M(F(x, x^n))$ converges to
the 2D Mahler measure $\log M(F(X, Y)) = \log\alpha_1$ as $n \to \infty$~\eqref{eq:exact_c_value}. To find how fast it converges and why it
oscillates around $\alpha_1$, we follow Boyd~\cite{boyd1981speculations} to
evaluate the leading term $\log\alpha_1$ as a 1D integral and subtract it from
$\log M(F(x, x^n))$ to isolate the error term.

First, we write $F(X, Y)$ as a quadratic in $Y$:
\begin{equation}
  \begin{aligned}
    F(X, Y) &= A_2(X)Y^2 + A_1(X)Y + A_0(X), \\
    A_2(X) &= X(X+1), \quad A_1(X) = X^2+X+1, \quad A_0(X) = X+1.
  \end{aligned}
  \label{eq:f_x_y_app_a2}
\end{equation}
By Vieta's formula, the two roots $Y_1(X)$ and $Y_2(X)$ satisfy $Y_1 Y_2 =
A_0(X)/A_2(X) = X^{-1}$ for $X \ne -1$. On the unit circle $|X| = 1$, their
product has modulus $|Y_1||Y_2| = 1$. Now ordering the roots so that $|Y_1| \ge
|Y_2|$ therefore gives $|Y_1| \ge 1 \ge |Y_2|$.

Next, we factor $F(X, e^{i\tau})$ and compute $\log|F(X, e^{i\tau})|$:
\begin{align}
  F(X, e^{i\tau}) &= A_2(X)(e^{i\tau} - Y_1)(e^{i\tau} - Y_2) =
  A_2(X)Y_1\left(\frac{e^{i\tau}}{Y_1} - 1\right)e^{i\tau}\left(1 -
  Y_2e^{-i\tau}\right), \nonumber \\
  |F(X, e^{i\tau})| &= |A_2(X) Y_1| \left|1 -
  \frac{e^{i\tau}}{Y_1}\right|\left|1 - \overline{Y_2}e^{i\tau}\right|,
  \nonumber \\
  \log|F(X, e^{i\tau})| &= \log|A_2(X) Y_1| + \log\bigl|1 -
  e^{i\tau}/Y_1\bigr| + \log\bigl|1 - \overline{Y_2}\,e^{i\tau}\bigr|.
  \label{eq:logabsF}
\end{align}
Since $|e^{i\tau}/Y_1| \le 1$ and $|\overline{Y_2}\,e^{i\tau}| \le 1$, we
expand the last two terms using $\log|1 - w| = -\frac{1}{2}\sum_{m=1}^{\infty}
\frac{1}{m}(w^m + \overline{w}^m)$ for $|w| \le 1$:
\begin{equation}
  \begin{aligned}
    \log\bigl|1 - e^{i\tau}/Y_1\bigr| &= -\frac{1}{2}\sum_{m=1}^{\infty}
    \frac{1}{m}\Bigl(Y_1^{-m} e^{im\tau} + \overline{Y_1}^{\,-m}
    e^{-im\tau}\Bigr),\\
    \log\bigl|1 - \overline{Y_2}\,e^{i\tau}\bigr| &=
    -\frac{1}{2}\sum_{m=1}^{\infty} \frac{1}{m}\Bigl(\overline{Y_2}^{\,m}
    e^{im\tau} + Y_2^m e^{-im\tau}\Bigr).
  \end{aligned}
\end{equation}
Adding the two sums gives the Fourier series $\sum_{m \ne 0} c_m(X)
e^{im\tau}$, with coefficients
\begin{equation}
  c_m(X) = -\frac{1}{2m}\Bigl(Y_1^{-m} + \overline{Y_2}^{\,m}\Bigr) \quad (m
  \ge 1), \qquad c_{-m}(X) = \overline{c_m(X)}.
  \label{eq:cm_def}
\end{equation}
Substituting this back into~\eqref{eq:logabsF} and computing the log Mahler measure~\eqref{eq:two_var_mahler_def} we get
\begin{equation}
  \log M(F(X, Y)) = \frac{1}{(2\pi)^2}\int_0^{2\pi}\!\!\int_0^{2\pi}
  \log\bigl|F(e^{it}, e^{i\tau})\bigr|\,d\tau\,dt = \frac{1}{2\pi}\int_0^{2\pi}
  \log\bigl|A_2(e^{it}) Y_1(e^{it})\bigr|\,dt.
  \label{eq:mahler_2d_a2}
\end{equation}
where we used $\int_0^{2\pi} e^{im\tau}\,d\tau = 0$ for every $m \ne 0$.

For the 1D Mahler measure $\log M(F(x, x^n))$ with $n = b/a \approx d/(2a)$,
setting $x = e^{it}$ locks $\tau = nt$ before integrating. So substituting $(X,
e^{i\tau}) = (e^{it}, e^{int})$ into~\eqref{eq:logabsF} and~\eqref{eq:cm_def}
gives
\begin{equation}
  \log\bigl|F(e^{it}, e^{int})\bigr| = \log\bigl|A_2(e^{it}) Y_1(e^{it})\bigr|
  + \sum_{m \ne 0} c_m(e^{it}) e^{imnt}.
\end{equation}
The log Mahler measure is then
\begin{equation}
    \log M(F(x, x^n)) = \frac{1}{2\pi}\int_0^{2\pi}\log\bigl|F(e^{it}, e^{int})\bigr| = \frac{1}{2\pi}\int_0^{2\pi}\log\bigl|A_2(e^{it}) Y_1(e^{it})\bigr|
  + \frac{1}{2\pi}\int_0^{2\pi}\sum_{m \ne 0} c_m(e^{it}) e^{imnt}.
  \label{eq:mahler_1d_a2}
\end{equation}
Finally we compute the desired difference between~\eqref{eq:mahler_1d_a2} and~\eqref{eq:mahler_2d_a2}:
\begin{equation}
  \log M(F(x, x^n)) - \log M(F(X, Y)) = \sum_{m \ne 0} \frac{1}{2\pi}
  \int_0^{2\pi} c_m(e^{it}) e^{imnt} \, dt.
  \label{eq:fourier_error_sum}
\end{equation}
where $c_m$ are functions of $Y_{1,2}$ given by~\eqref{eq:cm_def}.

To determine how $Y_{1,2}(e^{it})$ behave as a function of $t \in [-\pi, \pi]$,
we set $\tilde{Y} = e^{it/2} Y$ ($|\tilde{Y}| = |Y|$) and divide $F(e^{it}, Y)
= 0$ from~\eqref{eq:f_x_y_app_a2} by $e^{it/2}(e^{it} + 1)$, which gives a quadratic with real coefficients:
\begin{equation}
  \begin{aligned}
    \tilde{Y}^2 + B(t)\,\tilde{Y} + 1 &= 0, \qquad
    B(t) = \frac{2\cos t + 1}{2\cos(t/2)}, \\
    \tilde{Y}_{1,2}(t) &= \frac{-B(t) \pm \sqrt{B(t)^2 - 4}}{2}, \qquad
    B(t)^2 - 4 = \frac{4\cos^2 t - 4\cos t - 7}{4\cos^2(t/2)}.
  \end{aligned}
  \label{eq:rescaled_boyd_quadratic}
\end{equation}
Setting the numerator $4\cos^2 t - 4\cos t - 7 = 0$ gives the transition angle
\begin{equation}
  t_0 = \arccos\Bigl(\frac{1}{2} - \sqrt{2}\Bigr) \approx 2.7246\text{ rad}
  \quad (156.1^\circ).
  \label{eq:boyd_branch_angle}
\end{equation}
For $|t| < t_0$, $B(t)^2 - 4 < 0$, whereas for $|t| > t_0$, $B(t)^2 - 4 > 0$.
At $t = \pm t_0$, the numerator $4\cos^2 t - 4\cos t - 7$ has non-zero
derivative $(4 - 8\cos t_0)\sin(\pm t_0) = \pm 8\sqrt{2}\sin t_0 \ne 0$, and
the denominator $4\cos^2(t_0/2) = 3 - 2\sqrt{2}$ is positive. Thus $B(t)^2 - 4$
crosses zero linearly at $t = \pm t_0$, giving $\sqrt{B(t)^2 - 4} \propto |t
\mp t_0|^{1/2}$ for $|t| > t_0$.

Because $c_m(e^{it})$ in~\eqref{eq:cm_def} is built from $Y_{1,2}(e^{it})$, it
is smooth on $[-\pi, \pi]$ except at $t = \pm t_0$, where for $|t| > t_0$ it
has the square-root term $\gamma_m |t \mp t_0|^{1/2} + \mathcal{O}(|t \mp
t_0|^{3/2})$. Away from $t = \pm t_0$, integrating $e^{imnt}$ by parts $r$
times pulls out $(imn)^{-r}$ for any $r \ge 1$, so the smooth intervals decay
faster than any power of $1/n$. At $t = \pm t_0$, however, the derivative of
$|t \mp t_0|^{1/2}$ blows up like $|t \mp t_0|^{-1/2}$, stopping integration by
parts and leaving the neighborhoods of $t = \pm t_0$ as the dominant part
of~\eqref{eq:fourier_error_sum}. Substituting $u = |m|n(t - t_0)$ ($dt =
(|m|n)^{-1}\,du$) for $t > t_0$ pulls out $(|m|n)^{-1/2}$ from $(t -
t_0)^{1/2}$ and $(|m|n)^{-1}$ from $dt$:
\begin{equation}
  \int_{t_0}^{t_0 + \varepsilon} (t - t_0)^{1/2} e^{i m n t} \, dt = e^{i m n
  t_0} (|m|n)^{-3/2} \int_0^{|m|n\varepsilon} u^{1/2} e^{\pm i u} \, du \sim
  \mathcal{O}\!\left( n^{-3/2} \right).
\end{equation}
Adding the complex-conjugate contributions from $t = +t_0$ ($e^{+i m n t_0}$)
and $t = -t_0$ ($e^{-i m n t_0}$), substituting $n = b/a \approx d/(2a)$, and
exponentiating $\log M(P_d)$ gives the oscillatory asymptotic expansion:
\begin{equation}
  M(P_d) = \alpha_1 + A \cdot d^{-3/2} \cos(\omega d + \phi) +
  \mathcal{O}(d^{-5/2}),
  \label{eq:stationary_phase_law}
\end{equation}
where the $\mathcal{O}(d^{-5/2})$ remainder comes from the next $|t \mp
t_0|^{3/2}$ term. Whereas a fixed one-parameter sequence oscillates both above
and below $\alpha_1$, minimizing over $(a, b, k, s_0, s_1)$ at each even degree
$d$ selects parameters where $\cos(\omega d + \phi) \approx -1$, so the minimal
Mahler measure approaches $\alpha_1$ from below at rate $A \cdot d^{-3/2}$.

\subsection{Parameter Cutoff Analysis}
\label{app:dilation_saturation}

In Section~\ref{sec:analytic_resolution}, the Generalized Boyd
algorithm~\eqref{eq:generalized_boyd_5param} constructs polynomials $P_d(x) =
Q_{a,b,k,s_0,s_1}(x)/\Phi(x)$ with
\begin{equation}
  Q_{a,b,k,s_0,s_1}(x) = x^{a+k}(x^a + s_0)x^{2b}
    + s_1(x^{2a} + s_0 x^a + 1)x^b + s_0(x^a + s_0),
\label{eq:q_app_a3}
\end{equation}
where $a, b \in \mathbb{Z}^+$. Here $0 \le k < a$ is an exponent shift, $s_0, s_1
\in \{-1, +1\}$, and $\Phi(x)$ is the cyclotomic factor of degree $\delta =
\deg\Phi(x)$. We evaluated this family using a baseline cutoff $a \le 20$ and
an extended sweep up to $a < d/2$. A natural question is whether large values
of $a$ can yield asymptotically smaller Mahler measures. In this section, we
use a two-torus Fourier expansion to show that the lower envelope $\alpha_1 - A
\cdot d^{-3/2}$ is \emph{invariant} under $a$ for $a \ll \sqrt{d}$ and decays
once $a \gtrsim \sqrt{d}$.

Setting $k = 0$ (where all optimal reciprocal polynomials occur)
in~\eqref{eq:q_app_a3} gives the numerator $F(s_0 x^a, s_1 x^b)$ of degree $2a + 2b$,
so $P_d(x)$ has degree $d = 2a + 2b - \delta$. For $\gcd(a, b) = 1$ and
$\delta \ll d$, this gives $b \approx d/2 - a$.
Now to evaluate $\log M(F(s_0 x^a, s_1 x^b))$ we recall from~\eqref{eq:logabsF} and~\eqref{eq:cm_def} that on $|X| = |Y| = 1$,
\begin{equation}
  \log|F(X, Y)| = \log|A_2(X)Y_1(X)| - \operatorname{Re}\sum_{m=1}^{\infty}
  \frac{1}{m}\Bigl(Y_1(X)^{-m} + \overline{Y_2(X)}^{\,m}\Bigr) Y^m.
\end{equation}
Setting $(X, Y) = (s_0 e^{ia\theta}, s_1 e^{ib\theta})$ and averaging over
$\theta \in [0, 2\pi]$ gives
\begin{align}
    \log M\big(F(s_0 x^a, s_1 x^b)\big)
    &= \frac{1}{2\pi}\int_0^{2\pi} \log\bigl|F(s_0 e^{ia\theta}, s_1
    e^{ib\theta})\bigr|\,d\theta\nonumber \\
    &= \log\alpha_1 - \operatorname{Re}\sum_{m=1}^{\infty} \frac{s_1^m}{m}
    \frac{1}{2\pi}\int_0^{2\pi} \Bigl(Y_1(s_0 e^{ia\theta})^{-m} +
    \overline{Y_2(s_0 e^{ia\theta})}^{\,m}\Bigr) e^{imb\theta}\,d\theta,
    \label{eq:log_m_app_a3}
\end{align}
where the first term equals $\log\alpha_1$ by~\eqref{eq:mahler_2d_a2}.
Shifting $\theta \mapsto \theta + 2\pi/a$ leaves $s_0 e^{ia\theta}$ unchanged,
while $e^{imb\theta}$ picks up the phase $e^{2\pi i mb/a}$. Thus the integral
over $\theta \in [0, 2\pi]$ vanishes unless $a$ divides $mb$. Since $\gcd(a, b)
= 1$, this holds if and only if $m = ra$ for $r \in \mathbb{Z}^+$. Now when $m =
ra$, the integrand has period $2\pi/a$, so its average over $[0, 2\pi]$ equals
its average over one period $[0, 2\pi/a]$. Setting $t = a\theta$ for $s_0 = 1$
(or $t = a\theta + \pi$ for $s_0 = -1$) gives $e^{it} = s_0 e^{ia\theta}$ and
maps one period onto $t \in [-\pi, \pi]$, with $e^{irab\theta} =
(e^{ia\theta})^{rb} = s_0^{rb} e^{irbt}$ and $dt = a\,d\theta$. This gives
\begin{equation}
  \frac{1}{2\pi}\int_0^{2\pi} \Bigl(Y_1(s_0 e^{ia\theta})^{-ra} +
  \overline{Y_2(s_0 e^{ia\theta})}^{\,ra}\Bigr) e^{irab\theta}\,d\theta =
  s_0^{rb}\,\hat{Z}_{ra}(rb),
\end{equation}
where $\hat{Z}_m(n)$ is the single-turn integral
\begin{equation}
  \hat{Z}_m(n) = \frac{1}{2\pi} \int_{-\pi}^{\pi} \Bigl( Y_1(e^{it})^{-m} +
  \overline{Y_2(e^{it})}^{\,m} \Bigr) e^{i n t} \, dt.
  \label{eq:hat_z_mn}
\end{equation}
Since $s_1^{ra} s_0^{rb} = (s_1^a s_0^b)^r$ and $\hat{Z}_{ra}(rb)$ is real,
substituting $m = ra$ into the sum~\eqref{eq:log_m_app_a3} gives
\begin{equation}
  \log M\big(F(s_0 x^a, s_1 x^b)\big) = \log \alpha_1 - \sum_{r=1}^{\infty}
  \frac{(s_1^a s_0^b)^r}{r a} \, \hat{Z}_{r a}(r b).
  \label{eq:exact_2d_fourier_a}
\end{equation}
To evaluate $\hat{Z}_{ra}(rb)$ in~\eqref{eq:hat_z_mn}, we split the
integration variable $t \in [-\pi, \pi]$ at the branch angle $t_0$
~\eqref{eq:boyd_branch_angle}. On the exterior arc $|t| \in (t_0, \pi]$
($t = \pm t_0 \pm \tau$, $\tau > 0$), $2\cos t + 1 < 0$ so $B(t) < -2$
in~\eqref{eq:rescaled_boyd_quadratic}; the roots $\tilde{Y}_{1,2}(t) =
e^{\pm w(\tau)}$ are therefore positive real numbers with $w(\tau) = \kappa_0
\sqrt{\tau} + \mathcal{O}(\tau^{3/2}) > 0$, giving $Y_{1,2}(e^{it}) =
e^{-it/2} e^{\pm w(\tau)}$. Because~\eqref{eq:hat_z_mn} contains
$\overline{Y_2(e^{it})}$ rather than $Y_2(e^{it})$, conjugating
$Y_2(e^{it}) = e^{-it/2} e^{-w(\tau)}$ reverses its phase to $+t/2$ while
keeping its real modulus, so $\overline{Y_2(e^{it})} = e^{it/2} e^{-w(\tau)} =
Y_1(e^{it})^{-1}$. Both terms in~\eqref{eq:hat_z_mn} therefore equal
$e^{imt/2} e^{-mw(\tau)}$ and sum to $2 e^{i m t / 2} e^{-m w(\tau)}$.
Evaluating the $r = 1$ leading term $-\frac{1}{a} \hat{Z}_a(b)$ near $\tau = 0$
by comparing the decay factor $e^{-a \kappa_0 \sqrt{\tau}}$ against the
oscillation period $\tau \sim 2/d$ (via the ratio $\lambda = a \kappa_0
\sqrt{2/d}$) separates two regimes:
\begin{enumerate}
  \item \textbf{Regime I ($1 \le a \ll \sqrt{d}$, $\lambda \ll 1$): Exact
    Cancellation of $a$}. Because $e^{i(b + a/2)\tau} \approx e^{i(d/2)\tau}$
    oscillates on the scale $\tau \sim 2/d$, for $\lambda \ll 1$ we expand
    $e^{-a \kappa_0 \sqrt{\tau}} = 1 - a \kappa_0 \sqrt{\tau} +
    \mathcal{O}(a^2 \tau)$. The leading term $1$ matches the smooth unit-circle
    integrand $2e^{iat/2}$ on $|t| < t_0$ and integrates to zero over the full
    circle $[-\pi, \pi]$, whereas the one-sided $\sqrt{\tau}$ branch cut on
    $|t| > t_0$ integrates to $\hat{Z}_a(b) \sim C_0 \, a \, (d/2)^{-3/2}
    \cos(\omega d + \phi_a)$. Substituting this into~\eqref{eq:exact_2d_fourier_a}, the factor of $a$ cancels the
    $\frac{1}{a}$ prefactor:
    \begin{equation}
      \Delta M(d; a) \approx -\alpha_1 \frac{s_1^a s_0^b}{a} \hat{Z}_a(b) \sim
      -A \cdot d^{-3/2} \cos(\omega d + \phi_a).
      \label{eq:a_cancellation_law}
    \end{equation}
    Consequently, increasing $a$ within $1 \le a \ll \sqrt{d}$ does not change
    the amplitude $A \cdot d^{-3/2}$; it only varies the discrete phase
    $\phi_a$ and cyclotomic divisibility conditions so that $\cos(\omega d +
    \phi_a) \approx 1$ can be reached.
  \item \textbf{Regime II ($a \gtrsim \sqrt{d}$, $\lambda \gtrsim 1$):
    Exponential Damping}. Once $a \gtrsim \sqrt{d}$, the factor $e^{-a \kappa_0
    \sqrt{\tau}}$ decays within a single oscillation period $\tau \sim 2/d$,
    suppressing the integral by $\mathcal{O}(\lambda^{-3}) =
    \mathcal{O}((d/a^2)^{3/2})$ for $a \gg \sqrt{d}$. Thus, large $a$ values
    produce strictly smaller reductions below $\alpha_1$.
\end{enumerate}

Direct numerical evaluation across $a_{\max} \in \{20, 50, 100, d/9, d/2\}$
confirms both regimes: for $d \le 2000$, the optimal $a$ over $1 \le a < d/2$
always occurs at moderate $a \lesssim 2\sqrt{d} \le 90$, with zero improvement
between $a = 100$ and $a = d/2$. Furthermore, because the inner shift $k \in
\{0, \dots, a-1\}$ in~\eqref{eq:generalized_boyd_5param} already provides
dense phase sampling for $a \le 20$, expanding the search cutoff beyond $a \le
20$ leaves the lower envelope unchanged and refines only discrete phase
alignment.\footnote{Expanding the cutoff from $a \le 20$ to $a \le d/9$ (or $a
< d/2$) yields a median refinement of only $6.4 \times 10^{-6}$ across all 995
degrees $d \in [12, 2000]$ ($2.8 \times 10^{-5}$, or $0.0022\%$, for $d \ge
500$).}

\section{Root Distribution Analysis}
\label{app:spectral_details}

\subsection{Exact Unit-Circle Root Counting via Sturm's Theorem}
\label{app:unit_circle_sturm}

In section~\ref{sec:spectral_distributions} we saw that about $82.5\%$ of the
roots across all $824$ benchmark polynomials
($d \le 2000$) lie directly on $|z| = 1$ (see (Figure~\ref{fig:wilkinson_sturm}).
To count this number
$N_{\mathrm{uc}}(P_d)$ without floating-point error, we use the reciprocal
symmetry $P(z) = z^{2m} P(1/z)$ ($d = 2m$). Dividing by $z^m$ and substituting
$w = z + z^{-1} = 2\cos\theta \in [-2, 2]$ maps the unit circle $|z| = 1$ onto
the real interval $[-2, 2]$:
\begin{equation}
  \frac{P(z)}{z^m} = a_m + \sum_{k=1}^m a_{m-k} \left( z^k + z^{-k} \right) =
  a_m + \sum_{k=1}^m a_{m-k} V_k(w) = Q(w) \in \mathbb{Z}[w],
  \label{eq:chebyshev_projection}
\end{equation}
where $V_k(w) = 2 T_k(w/2)$ is the rescaled Chebyshev
polynomial.\footnote{Defined by the integer recurrence $V_0(w) = 2$,
$V_1(w) = w$, and $V_{k+1}(w) = w V_k(w) - V_{k-1}(w)$.} Because $P(z)$ is
irreducible of degree $d \ge 2$, it has no repeated roots and cannot vanish at
$z = \pm 1$ (otherwise $z \mp 1$ would divide $P(z)$). Consequently, $Q(w)$
does not vanish at the endpoints $w = \pm 2$, and every real root $w_0 \in
(-2, 2)$ of $Q(w)$ is a simple root corresponding to a complex-conjugate pair
$e^{\pm i\arccos(w_0/2)}$ on $|z| = 1$. Applying Sturm's theorem over
$\mathbb{Q}$ to count the real roots of $Q(w)$ in $(-2, 2)$ therefore gives the
exact unit-circle root count:\footnote{Forming the signed remainder sequence
$f_0(w) = Q(w)$, $f_1(w) = Q'(w)$, and $f_{k+1}(w) =
-\operatorname{rem}(f_{k-1}, f_k)$ in exact rational arithmetic,
$V_{\text{sign}}(-2) - V_{\text{sign}}(2)$ counts the real roots of $Q(w)$ in
$(-2, 2)$.}
\begin{equation}
  N_{\mathrm{uc}}(P) = 2 \Big( V_{\text{sign}}(-2) - V_{\text{sign}}(2) \Big).
  \label{eq:exact_sturm_count}
\end{equation}
As shown in Figure~\ref{fig:wilkinson_sturm} and Table~\ref{tab:sturm_audit},
evaluating~\eqref{eq:exact_sturm_count} across all 824 benchmark
polynomials shows that the unit-circle root fraction $N_{\mathrm{uc}}(P_d)/d$
stays nearly constant across two orders of magnitude in degree, lying within
$[77.0\%, 88.0\%]$ with a global mean of $82.51\%$.\footnote{For pure
specializations $F(s_0 x^a, s_1 x^b)$ as $b/a \to \infty$, both $Y$-roots of
$F(e^{it}, Y) = 0$ lie on $|Y| = 1$ for $|t| < t_0 = \arccos(\frac{1}{2} -
\sqrt{2})$, explaining the upper ceiling $t_0/\pi \approx 86.72\%$ visible in
Figure~\ref{fig:wilkinson_sturm}(a).}

\begin{table}[t!]
  \centering
  \caption{\textbf{Exact Unit-Circle Root Statistics Across Degree Regimes}
  computed via exact root counting over $\mathbb{Q}$.}
  \label{tab:sturm_audit}
  \small
  \setlength{\tabcolsep}{0pt}
  \begin{tabular*}{\textwidth}{@{\extracolsep{\fill}}l r r r r r@{}}
    \toprule
    \textbf{Degree Regime} & \textbf{Polys} &
    \textbf{Total Roots} &
    \textbf{Unit-Circle $N_{\mathrm{uc}}$} &
    \textbf{Unweighted Mean $N_{\mathrm{uc}}/d$} &
    \textbf{Interdecile Range} \\
    \midrule
    Degrees $10\text{--}100$ & 46 & 2,530 & 2,020 & $79.84\%$ &
    $[75.0\%, 84.6\%]$ \\
    Degrees $102\text{--}300$ & 100 & 20,100 & 16,708 & $83.12\%$ &
    $[79.2\%, 86.4\%]$ \\
    Degrees $302\text{--}700$ & 199 & 99,700 & 82,212 & $82.46\%$ &
    $[78.5\%, 86.1\%]$ \\
    Degrees $702\text{--}1500$ & 306 & 336,800 & 278,130 & $82.58\%$ &
    $[78.8\%, 86.3\%]$ \\
    Degrees $1502\text{--}2000$ & 173 & 293,950 & 243,358 & $82.78\%$ &
    $[79.1\%, 86.5\%]$ \\
    \midrule
    \textbf{Global Ensemble} & \textbf{824} & \textbf{753,080} &
    \textbf{622,428} & $\mathbf{82.51\%}$ & $\mathbf{[77.0\%, 88.0\%]}$ \\
    \bottomrule
  \end{tabular*}
\end{table}

\subsection{Off-Circle Root Structure}
\label{app:off_circle_clusters}

\begin{figure}[t!]
  \centering
  \includegraphics[width=\textwidth]{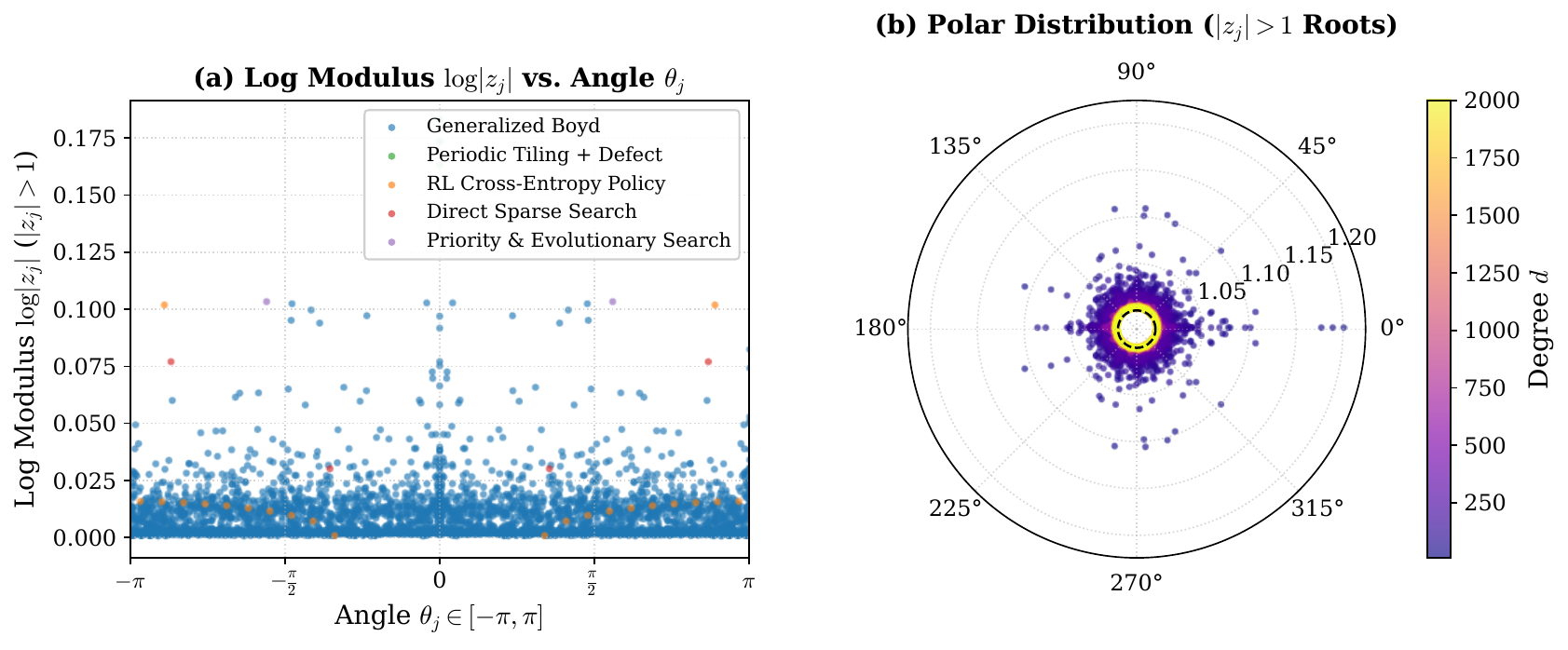}
  \caption{\textbf{Angular and radial distribution of off-circle roots ($|z_j|
  > 1$).} \textbf{(a)} Log modulus $\log|z_j|$ versus angle $\theta_j \in
  [-\pi, \pi]$ grouped by discovery method, showing radial contraction
  $\log|z_j| = \mathcal{O}(1/d)$ across all angles. \textbf{(b)} Polar view of
  exterior roots ($|z_j| > 1$) around the unit circle $|z|=1$, colored by
  degree $d$. Note that by reciprocal symmetry, every exterior root at $z_j$ is
  paired with an interior root at $1/\overline{z_j}$ at the same angle
  $\theta_j$ with $\log|1/\overline{z_j}| = -\log|z_j|$.}
  \label{fig:off_circle}
\end{figure}

While Section~\ref{sec:spectral_distributions} showed that about $82.5\%$ of
the roots lie directly on $|z| = 1$, at any degree $d$ the remaining $\approx
17.5\%$ of roots sit off the unit circle in reciprocal pairs ($|z_j| > 1$ and
$|z_j|^{-1} < 1$, Figure~\ref{fig:constellations}). Because the number of
exterior roots grows proportionally to $d$ while their positive logarithms
$\log|z_j|$ must sum to $\log M(P_d) < \log\alpha_1 \approx 0.2275$, their
radial distance from the unit circle shrinks at rate
$\log|z_j| = \mathcal{O}(1/d)$ (Figure~\ref{fig:off_circle}), consistent with
the Schinzel--Zassenhaus
theorem~\cite{schinzel1965conjecture,dimitrov2019proof}.\footnote{For example,
at $d = 1846$ every exterior root satisfies $\max_j |z_j| - 1 \le 0.003$.}

To see where these off-circle roots appear around the circle, recall
from~\eqref{eq:rescaled_boyd_quadratic} and~\eqref{eq:boyd_branch_angle} that
for $X = e^{it}$, both $Y$-roots of $F(e^{it}, Y) = 0$ lie on the unit circle
$|Y_{1,2}(e^{it})| = 1$ when $|t| < t_0 \approx 156.1^\circ$, and split
radially into $|Y_1(e^{it})| > 1 > |Y_2(e^{it})| = |Y_1(e^{it})|^{-1}$ on the
complementary arc $|t| > t_0$ near $t = \pi$. Under the one-variable
specialization $(X, Y) = (s_0 x^a, s_1 x^b)$, roots of $P_d(x)$ lift off $|z| =
1$ only along the arcs where $X = s_0 x^a$ passes through $|t| > t_0$, giving
two regimes:
\begin{itemize}
  \item \textbf{Single Cluster}: When $a = 1$, $X = \pm x$ winds once
    around the unit circle, so all off-circle roots gather in a single arc
    near $\theta = \pi$ (for $s_0 = +1$) or $\theta = 0$ (for $s_0 =
    -1$).\footnote{For instance, in the one-parameter degree-600 polynomial
    $P_{600}$~\eqref{eq:deg600_telescope}, all 40 exterior/interior root
    pairs lie within $|\theta - \pi| < 23^\circ$ ($\operatorname{Re}(z) <
    -0.92$).}
  \item \textbf{Symmetric Clusters}: When $a > 1$, the map $X =
    s_0 x^a$ winds $a$ times around the circle, pulling the off-circle arc
    $|t| > t_0$ back to $a$ evenly spaced clusters separated by angle $2\pi/a$
    around $|z| = 1$.\footnote{For instance, $P_{300}(x) = F(x^{13}, -x^{137})$
    ($a = 13$) has $13$ evenly spaced off-circle clusters around $|z| = 1$, and
    similarly $a = 5$ and $a = 31$ at $d = 100$ and $d = 1846$ in
    Figure~\ref{fig:constellations}.}
\end{itemize}

\bibliography{main}

@article{lehmer1933factorization,
  title={Factorization of certain cyclotomic functions},
  author={Lehmer, Derrick H.},
  journal={Annals of Mathematics},
  volume={34},
  number={3},
  pages={461--479},
  year={1933},
  publisher={JSTOR}
}

@article{smyth1971product,
  title={On the product of the conjugates outside the unit circle of an algebraic integer},
  author={Smyth, Christopher J.},
  journal={Bulletin of the London Mathematical Society},
  volume={3},
  number={2},
  pages={169--175},
  year={1971},
  publisher={Oxford University Press}
}

@article{dobrowolski1979question,
  title={On a question of {L}ehmer and the number of irreducible factors of a polynomial},
  author={Dobrowolski, Edward},
  journal={Acta Arithmetica},
  volume={34},
  number={4},
  pages={391--401},
  year={1979}
}

@article{cantor1982conjecture,
  title={On a conjecture of {D}.~{H}.~{L}ehmer},
  author={Cantor, David G. and Straus, Ernst G.},
  journal={Acta Arithmetica},
  volume={42},
  number={1},
  pages={97--100},
  year={1982}
}

@article{boyd1981speculations,
  title={Speculations concerning the range of {M}ahler's measure},
  author={Boyd, David W.},
  journal={Canadian Mathematical Bulletin},
  volume={24},
  number={4},
  pages={453--469},
  year={1981},
  publisher={Cambridge University Press}
}

@article{lawton1983problem,
  title={A problem of {B}oyd concerning geometric means of polynomials},
  author={Lawton, Wayne M.},
  journal={Journal of Number Theory},
  volume={16},
  number={3},
  pages={356--362},
  year={1983},
  publisher={Elsevier}
}

@article{boyd1998mahler,
  title={Mahler's measure and special values of {L}-functions},
  author={Boyd, David W.},
  journal={Experimental Mathematics},
  volume={7},
  number={1},
  pages={37--82},
  year={1998},
  publisher={Taylor \& Francis}
}

@article{deninger1997deligne,
  title={Deligne periods of mixed motives, {K}-theory and the entropy of certain $\mathbb{Z}^n$-actions},
  author={Deninger, Christopher},
  journal={Journal of the American Mathematical Society},
  volume={10},
  number={2},
  pages={259--281},
  year={1997}
}

@incollection{rodriguez1999modular,
  title={Modular {M}ahler measures {I}},
  author={Rodriguez-Villegas, Fernando},
  booktitle={Topics in Number Theory (University Park, PA, 1997)},
  series={Mathematics and Its Applications},
  volume={467},
  pages={17--48},
  year={1999},
  publisher={Kluwer Academic Publishers}
}

@incollection{kontsevich2001periods,
  title={Periods},
  author={Kontsevich, Maxim and Zagier, Don},
  booktitle={Mathematics Unlimited --- 2001 and Beyond},
  pages={771--808},
  year={2001},
  publisher={Springer}
}

@article{boyd2005small,
  title={Small limit points of {M}ahler's measure},
  author={Boyd, David W. and Mossinghoff, Michael J.},
  journal={Experimental Mathematics},
  volume={14},
  number={4},
  pages={403--414},
  year={2005},
  publisher={Taylor \& Francis}
}

@article{mossinghoff2008minimal,
  title={Minimal {M}ahler measures},
  author={Mossinghoff, Michael J. and Rhin, Georges and Wu, Qiang},
  journal={Experimental Mathematics},
  volume={17},
  number={4},
  pages={451--458},
  year={2008},
  publisher={Taylor \& Francis}
}

@article{schinzel1965conjecture,
  title={A refinement of two theorems of {K}ronecker},
  author={Schinzel, Andrzej and Zassenhaus, Hans},
  journal={Michigan Mathematical Journal},
  volume={12},
  number={1},
  pages={81--85},
  year={1965}
}

@article{dimitrov2019proof,
  title={A proof of the {S}chinzel--{Z}assenhaus conjecture on polynomials},
  author={Dimitrov, Vesselin},
  journal={arXiv preprint arXiv:1912.12545},
  year={2019}
}

@article{bilu1997equidistribution,
  title={Limit distribution of small points on algebraic tori},
  author={Bilu, Yuri},
  journal={Duke Mathematical Journal},
  volume={89},
  number={3},
  pages={465--476},
  year={1997}
}

@article{lind1990ergodic,
  title={Mahler measure and entropy for commuting automorphisms of compact groups},
  author={Lind, Douglas and Schmidt, Klaus and Ward, Tom},
  journal={Inventiones Mathematicae},
  volume={101},
  number={1},
  pages={593--629},
  year={1990}
}

@article{smyth1981mahler,
  title={On measures of polynomials in several variables},
  author={Smyth, Christopher J.},
  journal={Bulletin of the Australian Mathematical Society},
  volume={23},
  number={1},
  pages={49--63},
  year={1981}
}

@article{boyd1989reciprocal,
  title={Reciprocal polynomials having small measure. {II}},
  author={Boyd, David W.},
  journal={Mathematics of Computation},
  volume={53},
  number={187},
  pages={355--357},
  year={1989}
}

@article{mossinghoff1998list,
  title={Polynomials with small {M}ahler measure},
  author={Mossinghoff, Michael J.},
  journal={Mathematics of Computation},
  volume={67},
  number={224},
  pages={1697--1705},
  year={1998}
}

@article{davies2021advancing,
  title={Advancing mathematics by guiding human intuition with {AI}},
  author={Davies, Alex and Veli{\v{c}}kovi{\'c}, Petar and Buesing, Lars and Blackwell, Sam and Zheng, Daniel and Toma{\v{s}}ev, Nenad and Tanburn, Richard and Battaglia, Peter and Blundell, Charles and Juh{\'a}sz, Andr{\'a}s and Lackenby, Marc and Williamson, Geordie and Hassabis, Demis and Kohli, Pushmeet},
  journal={Nature},
  volume={600},
  number={7887},
  pages={70--74},
  year={2021}
}

@article{wagner2021constructions,
  title={Constructions in combinatorics via neural networks},
  author={Wagner, Adam Zsolt},
  journal={arXiv preprint arXiv:2104.14516},
  year={2021}
}

@article{fawzi2022discovering,
  title={Discovering faster matrix multiplication algorithms with reinforcement learning},
  author={Fawzi, Alhussein and Balog, Matej and Huang, Aja and Hubert, Thomas and Romera-Paredes, Bernardino and Barekatain, Mohammadamin and Novikov, Alexander and R.~Ruiz, Francisco J. and Schrittwieser, Julian and Swirszcz, Grzegorz and Silver, David and Hassabis, Demis and Kohli, Pushmeet},
  journal={Nature},
  volume={610},
  number={7930},
  pages={47--53},
  year={2022}
}

@article{romera2024mathematical,
  title={Mathematical discoveries from program search with large language models},
  author={Romera-Paredes, Bernardino and Barekatain, Mohammadamin and Novikov, Alexander and Balog, Matej and Kumar, M.~Pawan and Dupont, Emilien and Ruiz, Francisco J.~R. and Ellenberg, Jordan S. and Wang, Pengming and Fawzi, Omar and Kohli, Pushmeet and Fawzi, Alhussein},
  journal={Nature},
  volume={625},
  number={7995},
  pages={468--475},
  year={2024}
}

@article{trinh2024solving,
  title={Solving olympiad geometry without human demonstrations},
  author={Trinh, Trieu H. and Wu, Yuhuai and Le, Quoc V. and He, He and Luong, Thang},
  journal={Nature},
  volume={625},
  number={7995},
  pages={476--482},
  year={2024}
}

@article{charton2024patternboost,
  title={{PatternBoost}: Constructions in mathematics with a little help from {AI}},
  author={Charton, Fran{\c{c}}ois and Ellenberg, Jordan S. and Wagner, Adam Zsolt and Williamson, Geordie},
  journal={arXiv preprint arXiv:2411.00566},
  year={2024}
}

@article{novikov2025alphaevolve,
  title={{AlphaEvolve}: A coding agent for scientific and algorithmic discovery},
  author={Novikov, Alexander and Vu, Ngân and Eisenberger, Marvin and Dupont, Emilien and Huang, Po-Sen and Wagner, Adam Zsolt and Shirobokov, Sergey and Kozlovskii, Borislav and Ruiz, Francisco J.~R. and Mehrabian, Abbas and Kumar, M.~Pawan and See, Abigail and Chaudhuri, Swarat and Holland, George and Davies, Alex and Nowozin, Sebastian and Kohli, Pushmeet and Balog, Matej},
  journal={arXiv preprint arXiv:2506.13131},
  year={2025}
}

@article{shehper2024what,
  title={What makes math problems hard for reinforcement learning: a case study},
  author={Shehper, Ali and Medina-Mardones, Anibal M. and Fagan, Lucas and Lewandowski, Bart{\l}omiej and Gruen, Angus and Qiu, Yang and Kucharski, Piotr and Wang, Zhenghan and Gukov, Sergei},
  journal={Advances in Neural Information Processing Systems},
  volume={38},
  pages={177669--177702},
  year={2025}
}

@misc{gemini4argon2026,
  title={{Gemini 4 Argon}},
  author={{Google DeepMind}},
  year={2026},
  howpublished={\url{https://blog.google/innovation-and-ai/models-and-research/gemini-models/gemini-4-argon/}}
}

@misc{antigravity2026,
  title={{Google Antigravity}},
  author={{Google}},
  year={2026},
  howpublished={\url{https://antigravity.google/}}
}

@article{chervonyi2025gold,
  title={Gold-medalist performance in solving olympiad geometry with {AlphaGeometry2}},
  author={Chervonyi, Yuri and Trinh, Trieu H. and Ol{\v{s}}{\'a}k, Miroslav and Yang, Xiaomeng and Nguyen, Hoang and Menegali, Marcelo and Jung, Junehyuk and Kim, Junsu and Verma, Vikas and Le, Quoc V. and Luong, Thang},
  journal={arXiv preprint arXiv:2502.03544},
  year={2025}
}

@article{gottweis2025coscientist,
  title={Accelerating scientific discovery with {Co-Scientist}},
  author={Gottweis, Juraj and Weng, Wei-Hung and Daryin, Alexander and Tu, Tao and Sirkovic, Petar and Myaskovsky, Artiom and Glowaty, Grzegorz and Weissenberger, Felix and Orlandi, Alessio and Popovici, Dan and Palepu, Anil and Rong, Keran and Tanno, Ryutaro and Saab, Khaled and Zhang, Fan and Blum, Jacob and Carroll, Andrew and Kulkarni, Kavita and Toma{\v{s}}ev, Nenad and Zverinski, Dina and Rendulic, Ivor and Vedadi, Elahe and Hasler, Florian and Rimanic, Luka and Boia, Marina and Budiselic, Ivan and Feinstein, Ben and Bellaiche, Mathias and Sheffer, Tom and Freyberg, Jan and Ratcliff, Jeremy and Bertolli, Ottavia and Chou, Katherine and Hassidim, Avinatan and Gokturk, Burak and Vahdat, Amin and Guan, Yuan and Dhillon, Vikram and Vaishnav, Eeshit Dhaval and Lee, Byron and Costa, Tiago R.~D. and Penad{\'e}s, Jos{\'e} R. and Peltz, Gary and Matias, Yossi and Manyika, James and Hassabis, Demis and Xu, Yunhan and Kohli, Pushmeet and Pawlosky, Annalisa and Karthikesalingam, Alan and Natarajan, Vivek},
  journal={Nature},
  volume={655},
  number={8122},
  pages={487--496},
  year={2026}
}

@article{xia2026agentic,
  title={Agentic neurosymbolic collaboration for mathematical discovery: A case study in combinatorial design},
  author={Xia, Hai and Gomes, Carla P. and Selman, Bart and Szeider, Stefan},
  journal={arXiv preprint arXiv:2603.08322},
  year={2026}
}

@article{silver2016mastering,
  title={Mastering the game of {G}o with deep neural networks and tree search},
  author={Silver, David and Huang, Aja and Maddison, Chris J. and Guez, Arthur and Sifre, Laurent and Van Den Driessche, George and Schrittwieser, Julian and Antonoglou, Ioannis and Panneershelvam, Veda and Lanctot, Marc and Dieleman, Sander and Grewe, Dominik and Nham, John and Kalchbrenner, Nal and Sutskever, Ilya and Lillicrap, Timothy and Leach, Madeleine and Kavukcuoglu, Koray and Graepel, Thore and Hassabis, Demis},
  journal={Nature},
  volume={529},
  number={7587},
  pages={484--489},
  year={2016}
}

@article{lenstra1982factoring,
  title={Factoring polynomials with rational coefficients},
  author={Lenstra, Arjen K. and Lenstra, Hendrik W. and Lov{\'a}sz, L{\'a}szl{\'o}},
  journal={Mathematische Annalen},
  volume={261},
  number={4},
  pages={515--534},
  year={1982}
}

@article{boiko2023autonomous,
	title={Autonomous chemical research with large language models},
	author={Boiko, Daniil A. and MacKnight, Robert and Kline, Ben and Gomes, Gabe},
	journal={Nature},
	volume={624},
	number={7992},
	pages={570--578},
	year={2023},
	doi={10.1038/s41586-023-06792-0}
}

@article{swanson2025virtual,
	title={The {Virtual Lab} of {AI} agents designs new {SARS-CoV-2} nanobodies},
	author={Swanson, Kyle and Wu, Wesley and Bulaong, Nash L. and Pak, John E. and Zou, James},
	journal={Nature},
	volume={646},
	number={8085},
	pages={716--723},
	year={2025},
	doi={10.1038/s41586-025-09442-9}
}

@article{wang2023scientific,
	title={Scientific discovery in the age of artificial intelligence},
	author={Wang, Hanchen and Fu, Tianfan and Du, Yuanqi and Gao, Wenhao and Huang, Kexin and Liu, Ziming and Chandak, Payal and Liu, Shengchao and Van Katwyk, Peter and Deac, Andreea and Anandkumar, Anima and Bergen, Karianne and Gomes, Carla P. and Ho, Shirley and Kohli, Pushmeet and Lasenby, Joan and Leskovec, Jure and Liu, Tie-Yan and Manrai, Arjun and Marks, Debora and Ramsundar, Bharath and Song, Le and Sun, Jimeng and Tang, Jian and Veli{\v{c}}kovi{\'c}, Petar and Welling, Max and Zhang, Linfeng and Coley, Connor W. and Bengio, Yoshua and Zitnik, Marinka},
	journal={Nature},
	volume={620},
	number={7972},
	pages={47--60},
	year={2023}
}

@article{jumper2021highly,
	title={Highly accurate protein structure prediction with {AlphaFold}},
	author={Jumper, John and Evans, Richard and Pritzel, Alexander and Green, Tim and Figurnov, Michael and Ronneberger, Olaf and Tunyasuvunakool, Kathryn and Bates, Russ and {\v{Z}}{\'i}dek, Augustin and Potapenko, Anna and Bridgland, Alex and Meyer, Clemens and Kohl, Simon A. A. and Ballard, Andrew J. and Cowie, Andrew and Romera-Paredes, Bernardino and Nikolov, Stanislav and Jain, Rishub and Adler, Jonas and Back, Trevor and Petersen, Stig and Reiman, David and Clancy, Ellen and Zielinski, Michal and Steinegger, Martin and Pacholska, Michalina and Berghammer, Tamas and Bodenstein, Sebastian and Silver, David and Vinyals, Oriol and Senior, Andrew W. and Kavukcuoglu, Koray and Kohli, Pushmeet and Hassabis, Demis},
	journal={Nature},
	volume={596},
	number={7873},
	pages={583--589},
	year={2021}
}

@article{merchant2023scaling,
	title={Scaling deep learning for materials discovery},
	author={Merchant, Amil and Batzner, Simon and Schoenholz, Samuel S. and Aykol, Muratahan and Cheon, Gowoon and Cubuk, Ekin Dogus},
	journal={Nature},
	volume={624},
	number={7990},
	pages={80--85},
	year={2023}
}

@article{szymanski2023autonomous,
	title={An autonomous laboratory for the accelerated synthesis of novel materials},
	author={Szymanski, Nathan J. and Rendy, Bernardus and Fei, Yuxing and Kumar, Rishi E. and He, Tanjin and Milsted, David and McDermott, Matthew J. and Gallant, Max and Cubuk, Ekin Dogus and Merchant, Amil and Kim, Haegyeom and Jain, Anubhav and Bartel, Christopher J. and Persson, Kristin and Zeng, Yan and Ceder, Gerbrand},
	journal={Nature},
	volume={624},
	number={7990},
	pages={86--91},
	year={2023}
}

@article{degrave2022magnetic,
	title={Magnetic control of tokamak plasmas through deep reinforcement learning},
	author={Degrave, Jonas and Felici, Federico and Buchli, Jonas and Neunert, Michael and Tracey, Brendan and Carpanese, Francesco and Ewalds, Timo and Hafner, Roland and Abdolmaleki, Abbas and de las Casas, Diego and Donner, Craig and Fritz, Leslie and Galperti, Cristian and Huber, Andrea and Keeling, James and Tsimpoukelli, Maria and Kay, Jackie and Merle, Antoine and Moret, Jean-Marc and Noury, Seb and Pesamosca, Federico and Pfau, David and Sauter, Olivier and Sommariva, Cristian and Coda, Stefano and Duval, Basil and Fasoli, Ambrogio and Kohli, Pushmeet and Kavukcuoglu, Koray and Hassabis, Demis and Riedmiller, Martin},
	journal={Nature},
	volume={602},
	number={7897},
	pages={414--419},
	year={2022}
}

@article{lam2023learning,
	title={Learning skillful medium-range global weather forecasting},
	author={Lam, Remi and Sanchez-Gonzalez, Alvaro and Willson, Matthew and Wirnsberger, Peter and Fortunato, Meire and Alet, Ferran and Ravuri, Suman and Ewalds, Timo and Eaton-Rosen, Zach and Hu, Weihua and Merose, Alexander and Hoyer, Stephan and Holland, George and Vinyals, Oriol and Stott, Jacklynn and Pritzel, Alexander and Mohamed, Shakir and Battaglia, Peter},
	journal={Science},
	volume={382},
	number={6677},
	pages={1416--1421},
	year={2023}
}

@article{mirhoseini2021graph,
	title={A graph placement methodology for fast chip design},
	author={Mirhoseini, Azalia and Goldie, Anna and Yazgan, Mustafa and Jiang, Joe Wenjie and Songhori, Ebrahim and Wang, Shen and Lee, Young-Joon and Johnson, Eric and Pathak, Omkar and Nazi, Azade and Pak, Jiwoo and Tong, Andy and Srinivasa, Kavya and Hang, William and Tuncer, Emre and Le, Quoc V. and Laudon, James and Ho, Richard and Carpenter, Roger and Dean, Jeff},
	journal={Nature},
	volume={594},
	number={7862},
	pages={207--212},
	year={2021}
}

@article{stokes2020deep,
	title={A deep learning approach to antibiotic discovery},
	author={Stokes, Jonathan M. and Yang, Kevin and Swanson, Kyle and Jin, Wengong and Cubillos-Ruiz, Andres and Donghia, Nina M. and MacNair, Craig R. and French, Shawn and Carfrae, Lindsey A. and Bloom-Ackermann, Zohar and Tran, Victoria M. and Chiappino-Pepe, Anush and Badran, Ahmed H. and Andrews, Ian W. and Chory, Emma J. and Church, George M. and Brown, Eric D. and Jaakkola, Tommi S. and Barzilay, Regina and Collins, James J.},
	journal={Cell},
	volume={180},
	number={4},
	pages={688--702},
	year={2020}
}

@article{hubert2025olympiad,
	title={Olympiad-level formal mathematical reasoning with reinforcement learning},
	author={Hubert, Thomas and Mehta, Rishi and Sartran, Laurent and Horv{\'a}th, Mikl{\'o}s Z. and {\v{Z}}u{\v{z}}i{\'c}, Goran and Wieser, Eric and Huang, Aja and Schrittwieser, Julian and Schroecker, Yannick and Masoom, Hussain and Bertolli, Ottavia and Zahavy, Tom and Mandhane, Amol and Yung, Jessica and Beloshapka, Iuliya and Ibarz, Borja and Veeriah, Vivek and Yu, Lei and Nash, Oliver and Lezeau, Paul and Mercuri, Salvatore and S{\"o}nne, Calle and Mehta, Bhavik and Davies, Alex and Zheng, Daniel and Pedregosa, Fabian and Li, Yin and von Glehn, Ingrid and Rowland, Mark and Albanie, Samuel and Velingker, Ameya and Schmitt, Simon and others},
	journal={Nature},
	year={2025},
	doi={10.1038/s41586-025-09833-y}
}

@article{georgiev2025mathematical,
	title={Mathematical exploration and discovery at scale},
	author={Georgiev, Bogdan and G{\'o}mez-Serrano, Javier and Tao, Terence and Wagner, Adam Zsolt},
	journal={arXiv preprint arXiv:2511.02864},
	year={2025}
}

@article{Lind1982,
  author  = {Lind, D. A.},
  title   = {Dynamical properties of quasihyperbolic toral automorphisms},
  journal = {Ergodic Theory and Dynamical Systems},
  year    = {1982},
  volume  = {2},
  number  = {1},
  pages   = {49--68},
  doi     = {10.1017/S0143385700009573}
}

@article{LindSchmidtWard,
  author  = {Lind, Douglas and Schmidt, Klaus and Ward, Tom},
  title   = {{Mahler} measure and entropy for commuting automorphisms
             of compact groups},
  journal = {Inventiones Mathematicae},
  year    = {1990},
  volume  = {101},
  pages   = {593--629},
  doi     = {10.1007/BF01231517}
}

@article{hironaka2009what,
	title={What is\ldots {L}ehmer's number?},
	author={Hironaka, Eriko},
	journal={Notices of the American Mathematical Society},
	volume={56},
	number={3},
	pages={374--375},
	year={2009}
}

@article{smyth2015seventy,
	title={Seventy years of {S}alem numbers},
	author={Smyth, Christopher J.},
	journal={Bulletin of the London Mathematical Society},
	volume={47},
	number={3},
	year={2015}
}

@misc{wikiunsolved,
	title={List of unsolved problems in mathematics},
	author={{Wikipedia contributors}},
	year={2026},
	howpublished={\url{https://en.wikipedia.org/wiki/List_of_unsolved_problems_in_mathematics}},
	note={Section ``Analysis''; accessed October 2026}
}

@article{beilinson1985higher,
	title={Higher regulators and values of {L}-functions},
	author={Beilinson, Alexander A.},
	journal={Journal of Soviet Mathematics},
	volume={30},
	number={2},
	pages={2036--2070},
	year={1985}
}

@incollection{smyth2008mahler,
	title={The {M}ahler measure of algebraic numbers: a survey},
	author={Smyth, Christopher J.},
	booktitle={Number Theory and Polynomials},
	series={London Mathematical Society Lecture Note Series},
	volume={352},
	pages={322--349},
	year={2008},
	publisher={Cambridge University Press}
}

@article{farb2011small,
	title={Small dilatation pseudo-{A}nosov homeomorphisms and 3-manifolds},
	author={Farb, Benson and Leininger, Christopher J. and Margalit, Dan},
	journal={Advances in Mathematics},
	volume={228},
	number={3},
	pages={1466--1502},
	year={2011}
}

@article{mckee2012integer,
	title={Integer symmetric matrices of small spectral radius and small {M}ahler measure},
	author={McKee, James and Smyth, Christopher J.},
	journal={International Mathematics Research Notices},
	volume={2012},
	number={1},
	pages={102--136},
	year={2012}
}

@article{bergeron2013asymptotic,
	title={The asymptotic growth of torsion homology for arithmetic groups},
	author={Bergeron, Nicolas and Venkatesh, Akshay},
	journal={Journal of the Institute of Mathematics of Jussieu},
	volume={12},
	number={2},
	pages={391--447},
	year={2013}
}

@article{le2014homology,
	title={Homology torsion growth and {M}ahler measure},
	author={L{\^e}, Thang T. Q.},
	journal={Commentarii Mathematici Helvetici},
	volume={89},
	number={3},
	pages={719--757},
	year={2014}
}

@article{mcmullen2016automorphisms,
	title={Automorphisms of projective {K}3 surfaces with minimum entropy},
	author={McMullen, Curtis T.},
	journal={Inventiones Mathematicae},
	volume={203},
	number={1},
	pages={179--215},
	year={2016}
}

@article{emery2019salem,
	title={{S}alem numbers and arithmetic hyperbolic groups},
	author={Emery, Vincent and Ratcliffe, John G. and Tschantz, Steven T.},
	journal={Transactions of the American Mathematical Society},
	volume={372},
	number={1},
	pages={329--355},
	year={2019}
}

@book{brunault2020many,
	title={Many Variations of {M}ahler Measures: A Lasting Symphony},
	author={Brunault, Fran{\c{c}}ois and Zudilin, Wadim},
	series={Australian Mathematical Society Lecture Series},
	volume={28},
	publisher={Cambridge University Press},
	year={2020}
}

@book{mckee2021around,
	title={Around the Unit Circle: {M}ahler Measure, Integer Matrices and Roots of Unity},
	author={McKee, James and Smyth, Christopher J.},
	series={Universitext},
	publisher={Springer},
	address={Cham},
	year={2021}
}

@article{luck2022lehmer,
	title={{L}ehmer's problem for arbitrary groups},
	author={L{\"u}ck, Wolfgang},
	journal={Journal of Topology and Analysis},
	volume={14},
	number={4},
	pages={901--932},
	year={2022}
}

@article{ghate2001arithmetic,
	title={The arithmetic and geometry of {S}alem numbers},
	author={Ghate, Eknath and Hironaka, Eriko},
	journal={Bulletin of the American Mathematical Society},
	volume={38},
	number={3},
	pages={293--314},
	year={2001}
}

@misc{MathWorldProblems,
  author       = {Weisstein, Eric W.},
  title        = {Unsolved Problems},
  howpublished = {{MathWorld}---A Wolfram Resource},
  url          = {https://mathworld.wolfram.com/UnsolvedProblems.html},
  year         = {2026}
}

@book{bombieri2006heights,
	title={Heights in {D}iophantine Geometry},
	author={Bombieri, Enrico and Gubler, Walter},
	series={New Mathematical Monographs},
	volume={4},
	publisher={Cambridge University Press},
	year={2006}
}

@article{mcmullen2007dynamics,
	title={Dynamics on blowups of the projective plane},
	author={McMullen, Curtis T.},
	journal={Publications Math{\'e}matiques de l'IH{\'E}S},
	volume={105},
	pages={49--89},
	year={2007}
}

@article{yuzvinskii1967calculation,
	title={Calculation of the entropy of a group-endomorphism},
	author={Yuzvinski{\u\i}, S. A.},
	journal={Sibirski{\u\i} Matematicheski{\u\i} Zhurnal},
	volume={8},
	number={1},
	pages={230--239},
	year={1967},
	note={English translation: Siberian Mathematical Journal 8 (1967), no. 1, 172--178}
}

@book{schmidt1995dynamical,
	title={Dynamical Systems of Algebraic Origin},
	author={Schmidt, Klaus},
	series={Progress in Mathematics},
	volume={128},
	publisher={Birkh{\"a}user},
	address={Basel},
	year={1995}
}

@book{everest1999heights,
	title={Heights of Polynomials and Entropy in Algebraic Dynamics},
	author={Everest, Graham and Ward, Thomas},
	series={Universitext},
	publisher={Springer},
	year={1999}
}

@article{cantat2001dynamique,
	title={Dynamique des automorphismes des surfaces {K3}},
	author={Cantat, Serge},
	journal={Acta Mathematica},
	volume={187},
	number={1},
	pages={1--57},
	year={2001}
}

@article{hironaka2001lehmer,
	title={The {L}ehmer polynomial and pretzel links},
	author={Hironaka, Eriko},
	journal={Canadian Mathematical Bulletin},
	volume={44},
	pages={440--451},
	year={2001}
}

@article{mcmullen2002coxeter,
	title={{C}oxeter groups, {S}alem numbers and the {H}ilbert metric},
	author={McMullen, Curtis T.},
	journal={Publications Math{\'e}matiques de l'IH{\'E}S},
	volume={95},
	pages={151--183},
	year={2002}
}

@article{mcmullen2002dynamics,
	title={Dynamics on {K3} surfaces: {S}alem numbers and {S}iegel disks},
	author={McMullen, Curtis T.},
	journal={Journal f{\"u}r die reine und angewandte Mathematik},
	volume={545},
	pages={201--233},
	year={2002}
}

@article{riley1990growth,
	title={Growth of order of homology of cyclic branched covers of knots},
	author={Riley, Robert},
	journal={Bulletin of the London Mathematical Society},
	volume={22},
	number={3},
	pages={287--297},
	year={1990}
}

@article{gonzalezacuna1991cyclic,
	title={Cyclic branched coverings of knots and homology spheres},
	author={Gonz{\'a}lez-Acu{\~n}a, Francisco and Short, Hamish},
	journal={Revista Matem{\'a}tica de la Universidad Complutense de Madrid},
	volume={4},
	number={1},
	pages={97--120},
	year={1991}
}

@article{silver2002mahler,
	title={{M}ahler measure, links and homology growth},
	author={Silver, Daniel S. and Williams, Susan G.},
	journal={Topology},
	volume={41},
	number={5},
	pages={979--991},
	year={2002}
}

@book{luck2002l2,
	title={{$L^2$}-Invariants: Theory and Applications to Geometry and {K}-Theory},
	author={L{\"u}ck, Wolfgang},
	series={Ergebnisse der Mathematik und ihrer Grenzgebiete. 3. Folge},
	volume={44},
	publisher={Springer},
	year={2002}
}

@incollection{neumann1992arithmetic,
	title={Arithmetic of hyperbolic manifolds},
	author={Neumann, Walter D. and Reid, Alan W.},
	booktitle={Topology '90},
	series={Ohio State University Mathematical Research Institute Publications},
	volume={1},
	pages={273--310},
	year={1992},
	publisher={de Gruyter}
}

@book{maclachlan2003arithmetic,
	title={The Arithmetic of Hyperbolic 3-Manifolds},
	author={Maclachlan, Colin and Reid, Alan W.},
	series={Graduate Texts in Mathematics},
	volume={219},
	publisher={Springer},
	address={New York},
	year={2003}
}

@article{kenyon2006dimers,
	title={Dimers and amoebae},
	author={Kenyon, Richard and Okounkov, Andrei and Sheffield, Scott},
	journal={Annals of Mathematics},
	volume={163},
	number={3},
	pages={1019--1056},
	year={2006}
}

@incollection{boyd2002mahler,
	title={{M}ahler's measure and invariants of hyperbolic manifolds},
	author={Boyd, David W.},
	booktitle={Number Theory for the Millennium, {I}},
	pages={127--143},
	year={2002},
	publisher={A K Peters}
}

@article{gukov2005three,
	title={Three-dimensional quantum gravity, {C}hern--{S}imons theory, and the {A}-polynomial},
	author={Gukov, Sergei},
	journal={Communications in Mathematical Physics},
	volume={255},
	number={3},
	pages={577--627},
	year={2005}
}

@article{voutier1996effective,
	title={An effective lower bound for the height of algebraic numbers},
	author={Voutier, Paul},
	journal={Acta Arithmetica},
	volume={74},
	pages={81--95},
	year={1996}
}

@article{matveev2000explicit,
	title={An explicit lower bound for a homogeneous rational linear form in the logarithms of algebraic numbers. {II}},
	author={Matveev, E. M.},
	journal={Izvestiya: Mathematics},
	volume={64},
	number={6},
	pages={1217--1269},
	year={2000}
}

@article{friedman1989analytic,
	title={Analytic formulas for the regulator of a number field},
	author={Friedman, Eduardo},
	journal={Inventiones Mathematicae},
	volume={98},
	number={3},
	pages={599--622},
	year={1989}
}

@article{mignotte1974inequality,
	title={An inequality about factors of polynomials},
	author={Mignotte, Maurice},
	journal={Mathematics of Computation},
	volume={28},
	number={128},
	pages={1153--1157},
	year={1974}
}

@article{masser1989counting,
	title={Counting points of small height on elliptic curves},
	author={Masser, David W.},
	journal={Bulletin de la Soci{\'e}t{\'e} Math{\'e}matique de France},
	volume={117},
	pages={247--265},
	year={1989}
}

@article{amoroso1999probleme,
	title={Le probl{\`e}me de {L}ehmer en dimension sup{\'e}rieure},
	author={Amoroso, Francesco and David, Sinnou},
	journal={Journal f{\"u}r die reine und angewandte Mathematik},
	volume={513},
	pages={145--179},
	year={1999}
}

@article{david2000minoration,
	title={Minoration de la hauteur de {N}{\'e}ron--{T}ate sur les vari{\'e}t{\'e}s ab{\'e}liennes de type {C}.{M}.},
	author={David, Sinnou and Hindry, Marc},
	journal={Journal f{\"u}r die reine und angewandte Mathematik},
	volume={529},
	pages={1--74},
	year={2000}
}

@article{penner1991bounds,
	title={Bounds on least dilatations},
	author={Penner, Robert C.},
	journal={Proceedings of the American Mathematical Society},
	volume={113},
	number={2},
	pages={443--450},
	year={1991}
}

@article{leininger2004groups,
	title={On groups generated by two positive multi-twists: {T}eichm{\"u}ller curves and {L}ehmer's number},
	author={Leininger, Christopher J.},
	journal={Geometry \& Topology},
	volume={8},
	pages={1301--1359},
	year={2004}
}

@article{mcmullen2000polynomial,
	title={Polynomial invariants for fibered 3-manifolds and {T}eichm{\"u}ller geodesics for foliations},
	author={McMullen, Curtis T.},
	journal={Annales Scientifiques de l'{\'E}cole Normale Sup{\'e}rieure},
	volume={33},
	number={4},
	pages={519--560},
	year={2000}
}

@article{luck1999l2,
	title={{$L^2$}-torsion of hyperbolic manifolds of finite volume},
	author={L{\"u}ck, Wolfgang and Schick, Thomas},
	journal={Geometric and Functional Analysis},
	volume={9},
	pages={518--567},
	year={1999}
}

@article{lind2005lehmer,
	title={{L}ehmer's problem for compact abelian groups},
	author={Lind, Douglas},
	journal={Proceedings of the American Mathematical Society},
	volume={133},
	number={5},
	pages={1411--1416},
	year={2005}
}

@incollection{laurent1983minoration,
	title={Minoration de la hauteur de {N}{\'e}ron--{T}ate},
	author={Laurent, Michel},
	booktitle={S{\'e}minaire de Th{\'e}orie des Nombres, Paris 1981--82},
	series={Progress in Mathematics},
	volume={38},
	pages={137--151},
	year={1983},
	publisher={Birkh{\"a}user}
}

@article{call1993canonical,
	title={Canonical heights on varieties with morphisms},
	author={Call, Gregory S. and Silverman, Joseph H.},
	journal={Compositio Mathematica},
	volume={89},
	number={2},
	pages={163--205},
	year={1993}
}

@article{hughes2020neural,
	title={A neural network approach to predicting and computing knot invariants},
	author={Hughes, Mark C.},
	journal={Journal of Knot Theory and Its Ramifications},
	volume={29},
	number={3},
	pages={2050005},
	year={2020},
	note={arXiv:1610.05744 (2016)}
}

\end{document}